\documentclass[conference]{IEEEtran}
\IEEEoverridecommandlockouts
\usepackage{cite}
\usepackage{amsmath,amssymb,amsfonts}
\usepackage{algorithmic}
\usepackage{graphicx}
\usepackage{textcomp}
\usepackage{xcolor}
\def\BibTeX{{\rm B\kern-.05em{\sc i\kern-.025em b}\kern-.08em
		T\kern-.1667em\lower.7ex\hbox{E}\kern-.125emX}}

\usepackage{hyperref}
\usepackage{cleveref}
\usepackage{bm}
\usepackage{url}
\usepackage{layouts}
\usepackage{enumitem}
\usepackage{diagbox}
\usepackage{siunitx}
\usepackage{multirow}
\usepackage{hhline}
\usepackage{booktabs}
\usepackage{makecell}
\usepackage{float}
\usepackage{stfloats}
\usepackage{soul}
\usepackage[utf8]{inputenc}
\usepackage[small]{caption}
\usepackage{booktabs}
\usepackage{float}
\usepackage{caption}
\usepackage{subcaption}
\usepackage{csquotes}

\crefname{section}{Sec.}{Secs.}
\crefname{subsection}{Sec.}{Secs.}
\crefname{subsubsection}{Sec.}{Secs.}
\crefname{paragraph}{Par.}{Para.}
\crefname{appendix}{App.}{Apps.}
\crefname{table}{Tab.}{Tabs.}
\crefname{figure}{Fig.}{Figs.}
\crefname{equation}{Eq.}{Eq.}
\Crefname{equation}{Eq.}{Eq.}
\crefname{algorithm}{Alg.}{Algs.}

\newcommand{\mnist}[1]{\raisebox{-2px}{\includegraphics[width=14px,height=14px]{figs/mnist/#1}}}
\newcommand{\fmnist}[1]{\raisebox{-2px}{\includegraphics[width=14px,height=14px]{figs/fashionmnist/#1}}}
\newcommand{\devan}[1]{\raisebox{-2px}{\includegraphics[width=16px,height=16px]{figs/devanagari/#1}}}

\usepackage{comment}

\makeatletter
\def\ps@IEEEtitlepagestyle{%
	\def\@oddfoot{\mycopyrightnotice}%
	\def\@oddhead{\hbox{}\@IEEEheaderstyle\leftmark\hfil\thepage}\relax
	\def\@evenhead{\@IEEEheaderstyle\thepage\hfil\leftmark\hbox{}}\relax
	\def\@evenfoot{}%
}
\def\mycopyrightnotice{%
	\begin{minipage}{\textwidth}
		\centering \scriptsize
		Copyright~\copyright~2021 IEEE. Personal use of this material is permitted. Permission from IEEE must be obtained for all other uses, in any current or future media, including\\reprinting/republishing this material for advertising or promotional purposes, creating new collective works, for resale or redistribution to servers or lists, or reuse of any copyrighted component of this work in other works by sending a request to pubs-permissions@ieee.org.
	\end{minipage}
}
\makeatother

\begin{document}

\title{Overcoming Catastrophic Forgetting with\\Gaussian Mixture Replay}

\author{\IEEEauthorblockN{Benedikt Pf{\"u}lb}
\IEEEauthorblockA{\textit{Department of Applied Computer Science} \\
\textit{Fulda University of Applied Sciences}\\
Fulda, Germany \\
benedikt.pfuelb@cs.hs-fulda.de}
\and
\IEEEauthorblockN{Alexander Gepperth}
\IEEEauthorblockA{\textit{Department of Applied Computer Science} \\
\textit{Fulda University of Applied Sciences}\\
Fulda, Germany\\
alexander.gepperth@cs.hs-fulda.de}
}
\maketitle

\begin{abstract}
We present Gaussian Mixture Replay (GMR), a rehearsal-based approach for continual learning (CL) based on Gaussian Mixture Models (GMM). 
CL approaches are intended to tackle the problem of catastrophic forgetting (CF), which occurs for Deep Neural Networks (DNNs) when sequentially training them on successive sub-tasks.
GMR mitigates CF by generating samples from previous tasks and merging them with current training data.
GMMs serve several purposes here: sample generation, density estimation (e.g., for detecting outliers or recognizing task boundaries) and providing a high-level feature representation for classification.
GMR has several conceptual advantages over existing replay-based CL approaches.
First of all, GMR achieves sample generation, classification and density estimation in a single network structure with strongly reduced memory requirements.
Secondly, it can be trained at constant time complexity w.r.t.\ the number of sub-tasks, making it particularly suitable for life-long learning.
Furthermore, GMR minimizes a differentiable loss function and seems to avoid mode collapse.
In addition, task boundaries can be detected by applying GMM density estimation. 
Lastly, GMR does not require access to sub-tasks lying in the future for hyper-parameter tuning, allowing CL under real-world constraints.
We evaluate GMR on multiple image datasets, which are divided into class-disjoint sub-tasks.
\end{abstract}
\vspace{.5em}
\section{Introduction}
\noindent This article is set in the context of continual learning (CL).
Continual (or life-long) learning is an important element for successful biological systems.
CL denotes the ability to acquire new knowledge without forgetting previously learned knowledge.
The CL problem is usually formalized as sequentially training a model (e.g., a DNN) on data that is divided into a sequence of sub-tasks (here denoted as $T_1$, $T_2,\ \dots$). 
When doing this, a basic problem that occurs with DNNs is catastrophic forgetting (CF)~\cite{French1999}. 
\par
\begin{figure}[htb!]
	\centering
	\includegraphics[width=0.49\linewidth]{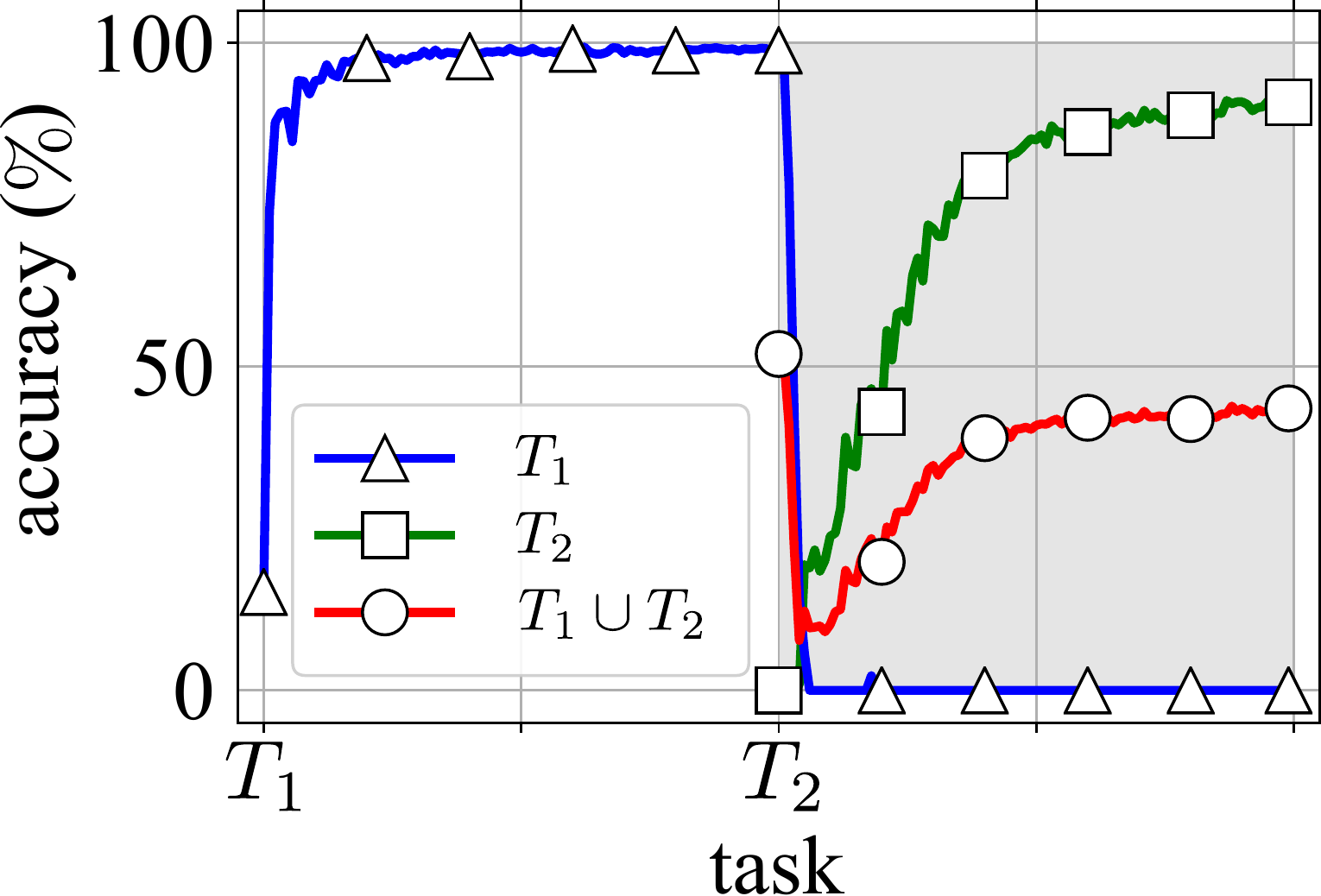}
	\includegraphics[width=0.49\linewidth]{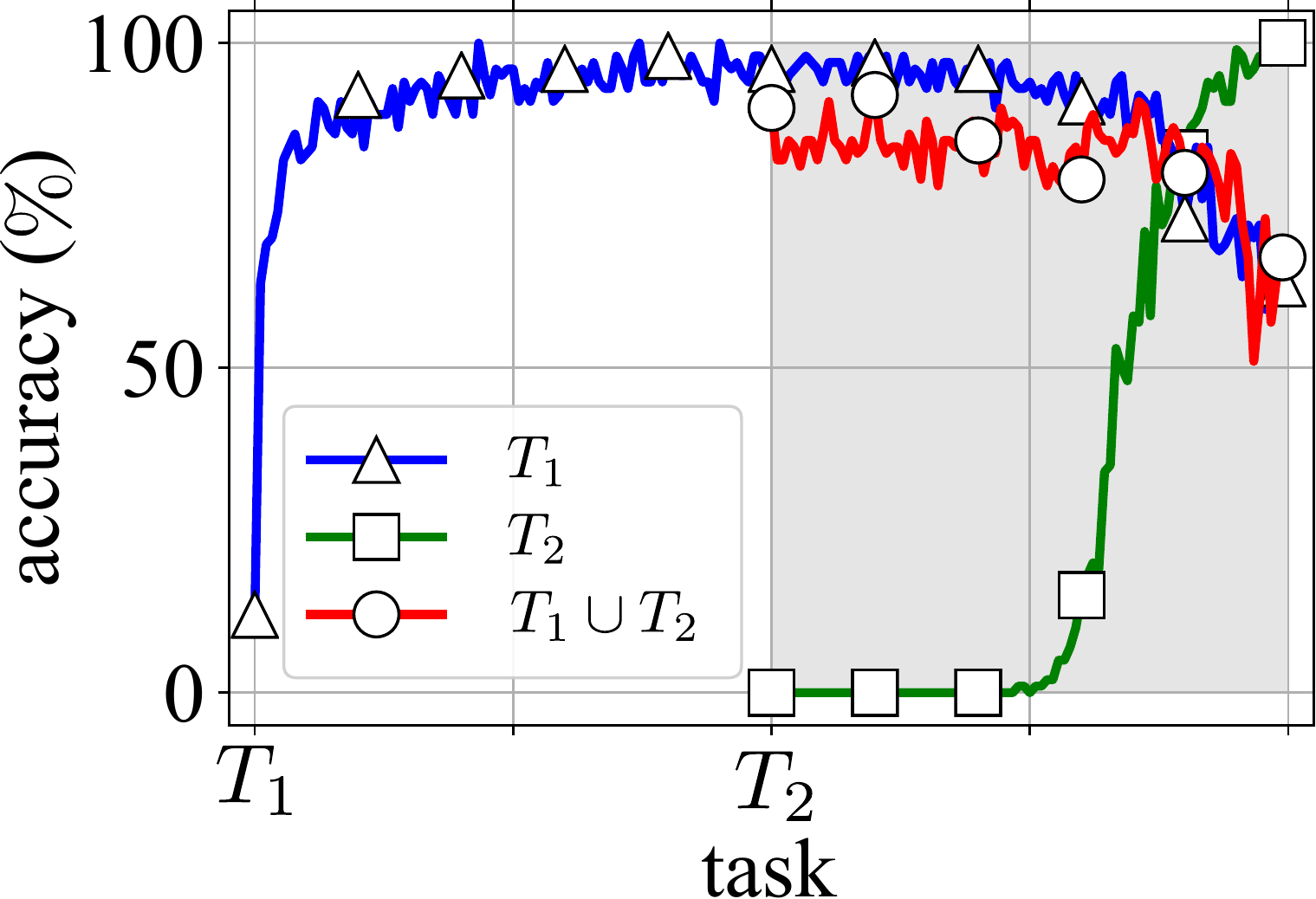}
	\caption{Visualization of the catastrophic forgetting (CF) effect~\cite{Pfuelb2019}. 
		Left image shows CF.  
		The blue line drops complete after a few training iterations on a second task.
		Hence, knowledge of previous task is lost.
		Right side shows a more controllable linear decline.
	}
	\label{fig:CF}
\end{figure}

\noindent\textbf{\mbox{Catastrophic Forgetting}} 
Essentially, the CF effect implies an abrupt and near-complete loss of knowledge about $T_1,$\,$\dots,$\,$T_{t-1}$ with a few training iterations on $T_t$ (see \cref{fig:CF}). 
\par
\noindent In real-world applications, many variations of the CL scenario can be distinguished.
This includes, for example, the problem of unavailable sub-task boundaries, which can even be fluid (concept drift/shift).
Many approaches (some of which are targeting very specific CL scenarios) for avoiding CF have been proposed recently, see \cref{sec:relwork}.
This article proposes a model that is based on \textit{replay}.
\par
\begin{figure}[htb!]
	\centering
	\includegraphics[width=\linewidth]{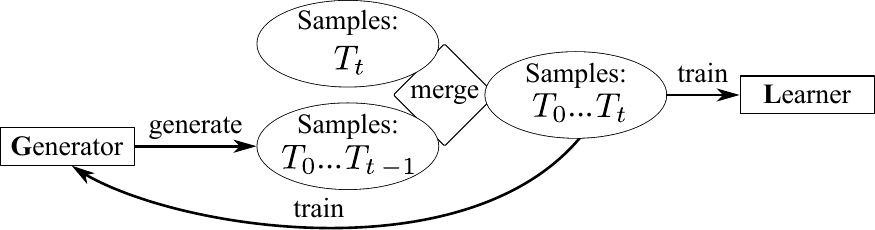}
	\caption{
		The replay approach to continual learning: a Learner (L), e.g., a DNN, is trained on several sub-tasks sequentially. 
		To avoid forgetting, a generator (G) is trained to generate samples from past sub-tasks.
		For training L, G \textit{generates} samples from past sub-tasks, which are merged with current sub-task samples.
	}
	\label{fig:replay}
\end{figure}

\noindent{\textbf{Replay}} 
An inspiration for avoiding CF is the learning process of students.
In school, new topics are continuously taught.
The challenge is to retain older topics when studying current ones.
In order to solve this problem, students must 1) recognize that they have forgotten something and 2) repeat and revise, if necessary.
This real-world association can serve as an inspiration for a replay approach towards CL, see \cref{fig:replay}.
\subsection{Gaussian Mixture Replay}
\begin{figure}[htb!]
\centering
\includegraphics[width=\linewidth]{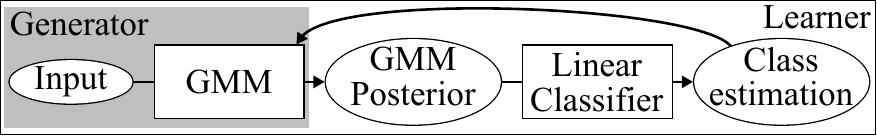}
\caption{
	Principal structure of the GMR model, composed of a GMM modeling the distribution of training samples (left), and a linear classifier operating on the posterior probabilities (sometimes termed \textit{responsibilities}) produced by the GMM.
	The coupled GMM/classifier implements the learner from \cref{fig:replay} (operating from left to right), whereas the GMM implements the generator from \cref{fig:replay}.
	The GMM sampling process is informed by feedback from the classifier (operating from right to left). 
	}
\label{fig:gmr}
\end{figure}

\noindent GMR is a new approach for lifelong learning based on Gaussian Mixture Models (GMMs).
GMMs aim to represent the distribution of training samples, enabling them to assess the probability of unknown samples (density estimation), as well as generate new samples from the learned distribution (sampling).
In the context of CL, density estimation is a useful tool for detecting task boundaries, because samples of a previously unseen nature (e.g., a new visual class) start arriving at task boundaries.
Furthermore, GMMs can play the role of a generator in replay-based CL due to their sampling ability see \cref{fig:replay} (e.g., sampling from a multivariate Gaussian distributions).
In GMR, the learner and the generator from \cref{fig:replay} are contained in a single structure.
The GMM layer provides sampling, density estimation, as well as a feature representation for the linear classifier layer.
This layer operates on quantities termed \textit{responsibilities} or posterior probabilities of the GMM layer, see \cref{fig:gmr}.
Responsibilities are normalized and lie in the range $[0\dots 1]$.
Therefore, they are ideal for linear classification.
In turn, by approximate inversion of the linear classifier layer, GMM sampling can be informed about the classes of samples that should be generated. 
\subsection{Related Work}\label{sec:relwork}
\noindent The field of CL is expanding rapidly, see \cite{Parisi2019, Hayes2018, Soltoggio2017, DeLange2019} for reviews.
Systematic comparisons between different approaches to avoid CF are performed in, e.g., \cite{Kemker2017,Pfuelb2019}.
As discussed in \cite{Pfuelb2019}, many recently proposed methods demand specific experimental setups, which deviate significantly from application scenarios.
For example, some methods require access to samples from \textit{all} sub-tasks for tuning hyper-parameters, whereas others need to store all samples from past tasks.
Many proposed methods have a time and/or memory complexity that scales at least linearly with the number of sub-tasks and, thus, may fail if this number is large.
Among the proposed remedies to CF, three major directions may be distinguished according to \cite{DeLange2019}: parameter isolation, regularization and replay.
\par
\noindent\textbf{\mbox{Parameter Isolation}} Isolation methods aim at determining (or creating) a group of DNN parameters that are mainly \enquote{responsible} for a certain sub-task. 
CL is then avoided by \textit{protecting} these parameters when training on successive sub-tasks. 
Representative works are \cite{Fernando2017,Mallya2017,Mallya2018,Serra2018,Rusu2016,Aljundi2016}.
\par
\noindent\textbf{Regularization} Regularization methods mostly propose modifying the DNN loss function, including additional terms that protect knowledge acquired in previous sub-tasks.
Current approaches are very diverse: SSL \cite{Aljundi2018} focuses on enhancing sparsity of neural activities, whereas approaches such as LwF \cite{Li2016} rely on knowledge distillation mechanisms.
A method that has attracted significant attention is Elastic Weight Consolidation (EWC) \cite{Kirkpatrick2016}, which inhibits changes to weights that are important to previous sub-tasks, measuring their importance based on the Fisher information matrix (FIM).
Synaptic intelligence \cite{Zenke2017} is pursuing a similar goal.
An online version of EWC is also published~\cite{Schwarz2018}.
Incremental Moment Matching (IMM) \cite{Lee2017} makes use of the FIM to merge the parameters obtained for different sub-tasks. 
The Matrix of Squares (MasQ) method \cite{Gepperth2019SIM} is identical to EWC, but relies on the calculus of derivatives to assess the importance of parameters for a given sub-task.
It is, therefore, more simple w.r.t\ its concepts and much more memory-efficient.
\par
\noindent\textbf{Replay}
Replay-based methods are at the core of this article.
They mainly occur in two forms: rehearsal and pseudo-rehearsal.
\textit{Rehearsal methods} store a subset of samples from past sub-tasks preventing CF, either by putting constraints on current sub-task training or by adding retained samples to the current sub-task training set.
Typical representatives of rehearsal methods are iCaRL \cite{Rebuffi2017a}, (A-)GEM \cite{LopezPaz2017, Chaudhry2018}, GBSS \cite{Aljundi2019} and TEM \cite{Chaudhry2019}. 
\textit{Pseudo-rehearsal} methods, in contrast, do not store samples but generate them using a dedicated \textit{generator} that is trained along with the learner, see \cref{fig:replay}.
Typical models used as generators are Generative Adversarial Networks (GANs), Variational Autoencoders (VAEs) and their variants, see \cite{Shin2017} and \cite{Kamra2017}.
While being impressive in terms of CL performance, these approaches have problems, as well.
First of all, training time scales linearly with the number of sub-tasks. 
Secondly, even for simple datasets like MNIST, the generators require significant computational resources, in addition to the resource requirements for the learner (CNNs or DNNs). 
Lastly, GANs can suffer from mode collapse, which is very hard to detect, since GANs do not minimize a differentiable loss function.
The GMR model we are proposing belongs to pseudo-reharsal approaches, as well.
\par
\noindent\textbf{Training and Evaluation Paradigms for CL}
In the context of continual learning, a variety of trainings and evaluation paradigms are proposed the literature \cite{Buzzega2020,Joseph2020,We2019,Farquhar2018,Kemker2017,Lesort2018}.
In addition to a very broad definition of various topics related to continuous learning, \cite{Lesort2020} addresses the topic of evaluation protocols and metrics.
The influence of different difficulties and the correlative order of tasks is addressed.
The evaluation of generative models is also highlighted.
De Lange et.\,al.\ describe the random selection of classes that are divided into tasks of size two~\cite{DeLange2019}.
It must be ensured that all tasks are approximately equal in difficulty.
Effectiveness is measured by determining the difference between the accuracy and the expected value after each task.
Another comprehensive survey is presented by \cite{Mundt2020}.
Here an evaluation strategy that is \enquote{more comprehensive} with regard to real-world applications is suggested, among other things.
The classical evaluation pipeline is extended by adding queries, e.g., related to the importance of the data.
In addition to human activities, statistical methods can be used to influence the training process. 
Parisi et\,al.\ \cite{Parisi2019} also address the issue of continual learning from a biological perspective.
Moreover, the benchmark procedure used is shown.
Again, \enquote{incremental class learning} is one of the methods used.
\subsection{Contributions}
\noindent GMR offers several novel contributions to the field of CL: 
\begin{itemize}[leftmargin=*,nosep]
	\setlength\itemsep{0em}
	\item Pseudo-rehearsal based CL integrating learner and generator in a single structure
	\item Easy-to-implement detection of task boundaries that is well-justified by probability theory
	\item Constant time complexity w.r.t.\ the number of sub-tasks
	\item No cross-validation on future sub-tasks for hyper-parameter selection required
\end{itemize}
To validate our approach, we perform a comparison to the Elastic Weight Consolidation (EWC) model which is viewed as a \enquote{standard model} for CL in many recent publications.
Furthermore, we provide a public TensorFlow implementation\footnote{\url{https://gitlab.cs.hs-fulda.de/ML-Projects/gmr}}.
\section{Datasets}\label{sec:datasets}
\noindent In order to measure the impact of forgetting during continual learning, three public datasets are used (see \cref{tab:datasets}).
All datasets consist of gray scale images with dimensions of $784$ or $1\,024$.
Each of them is divided in the ratio $90\%$ to $10\%$ in training and test data with an almost equal distribution of samples within classes.
The used datasets are normalized to the $[0,1]$ range.
\par
MNIST contains images of handwritten digits ($0$-$9$) with a resolution of $28$\,$\times$\,$28$ pixels.
It is probably the most commonly used benchmark for classification problems.
FashionMNIST contains pictures of different types of clothes.
This dataset is supposed to be harder to classify compared to MNIST (same resolution) and, thus, leads to lower accuracies.
Similar to MNIST, the Devanagari dataset contains written Devanagari letters. 
It is available in a resolution of $32$\,$\times$\,$32$ pixels per image.
Since there are more classes included than needed, we randomly select $10$ classes.
\begin{table*}[b]
	\newcolumntype{a}{>{\centering\arraybackslash}m{16px}}
	\centering
	\caption{Detailed information to the used datasets (including examples of each class).}
	\label{tab:datasets}
	\setlength\tabcolsep{2pt}
	\begin{tabular}{|l|c|c|c|c|aaaaaaaaaa|}
		\hline
		\multicolumn{1}{|c|}{\textbf{dataset}}  &   \textbf{ref.}    & \textbf{resolution}  &    \textbf{number of}     &  \textbf{number of}   &                                  \multicolumn{10}{c|}{random \textbf{examples} (from classes)}                                  \\
		                                        &                    &                      & \textbf{training samples} & \textbf{test samples} & 0          & 1          & 2          & 3          & 4          & 5          & 6          & 7          & 8          & 9          \\ \hline
		MNIST                                   &  \cite{Lecun1998}  & $28$\,$\times$\,$28$ &          50\,000          &        10\,000        & \mnist{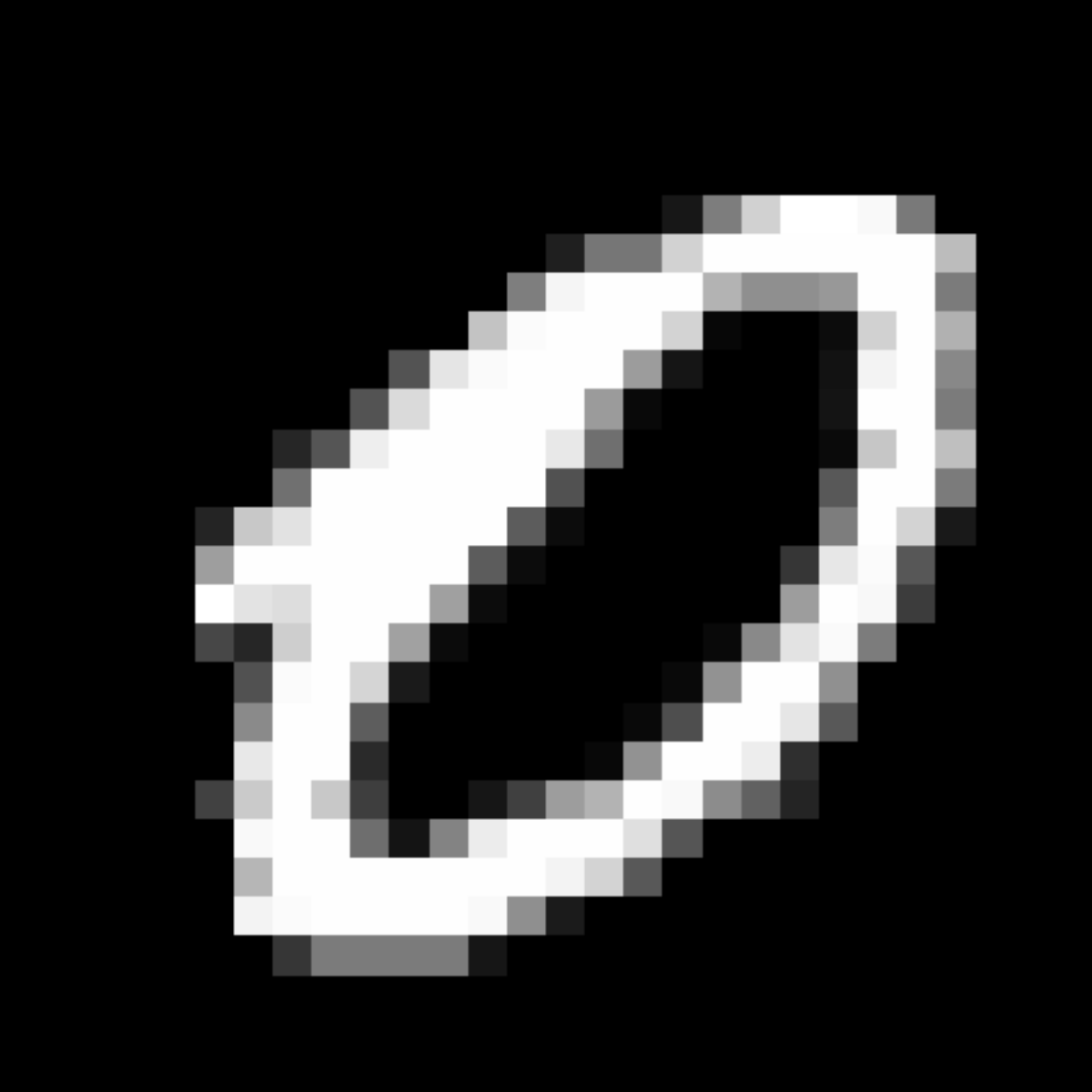}  & \mnist{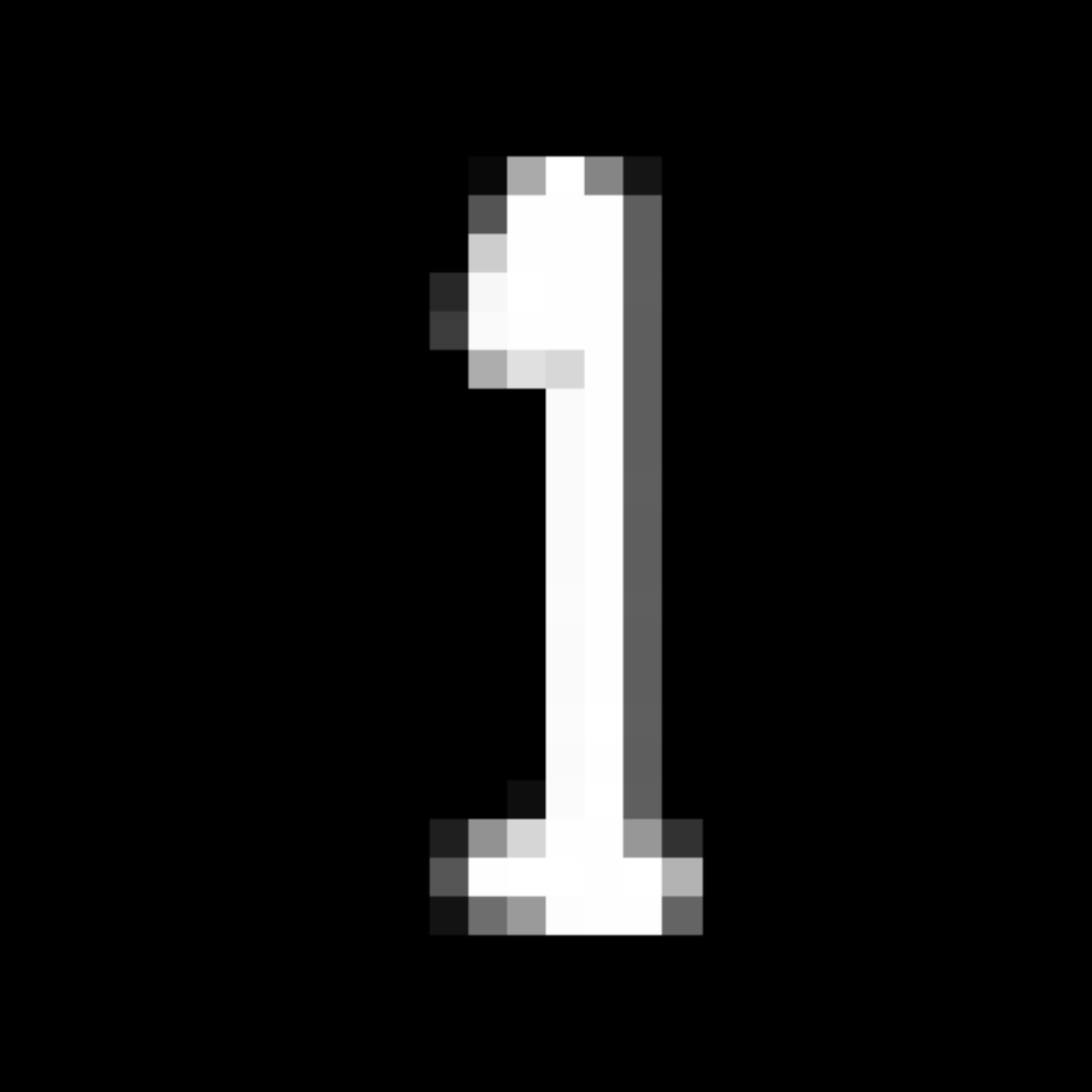}  & \mnist{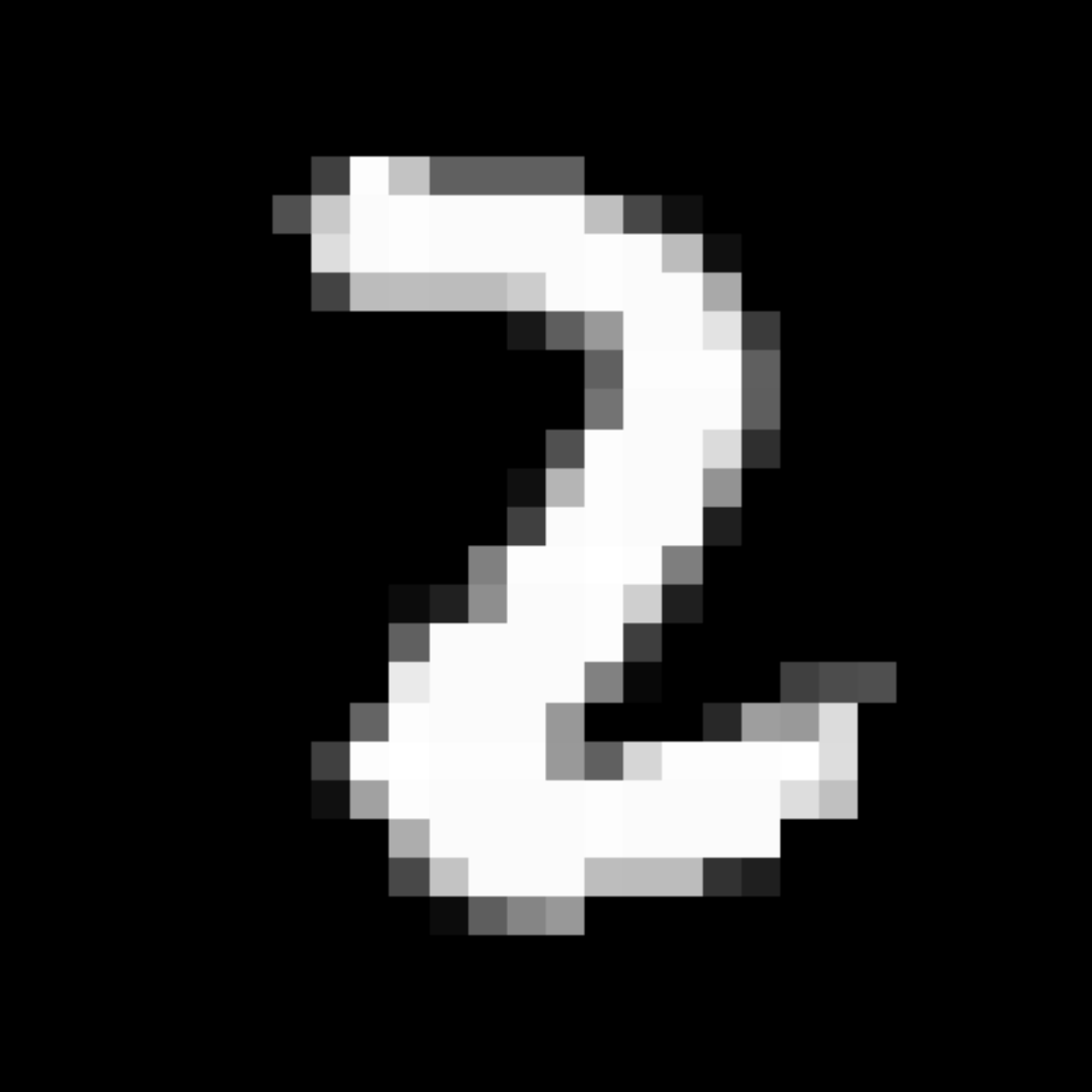}  & \mnist{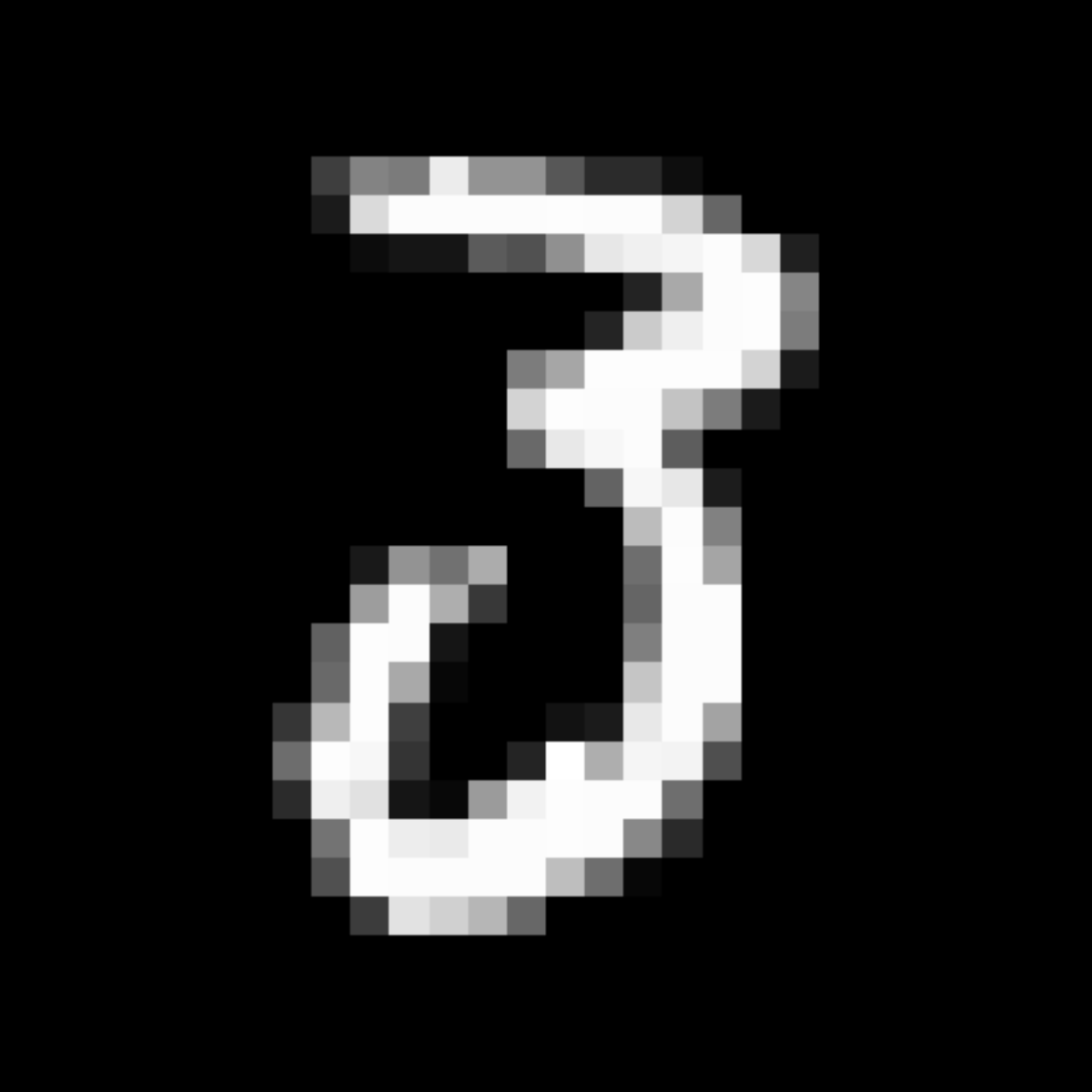}  & \mnist{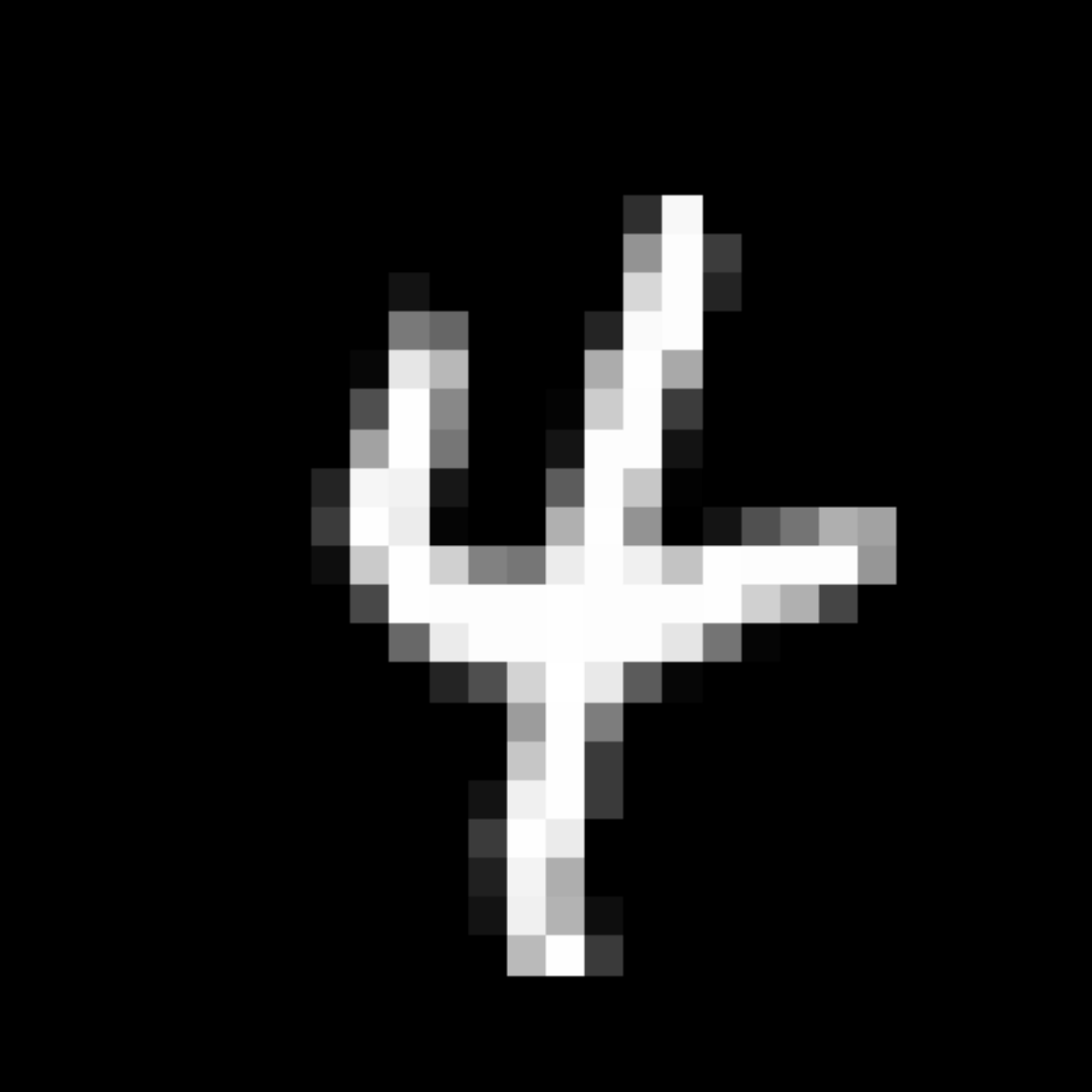}  & \mnist{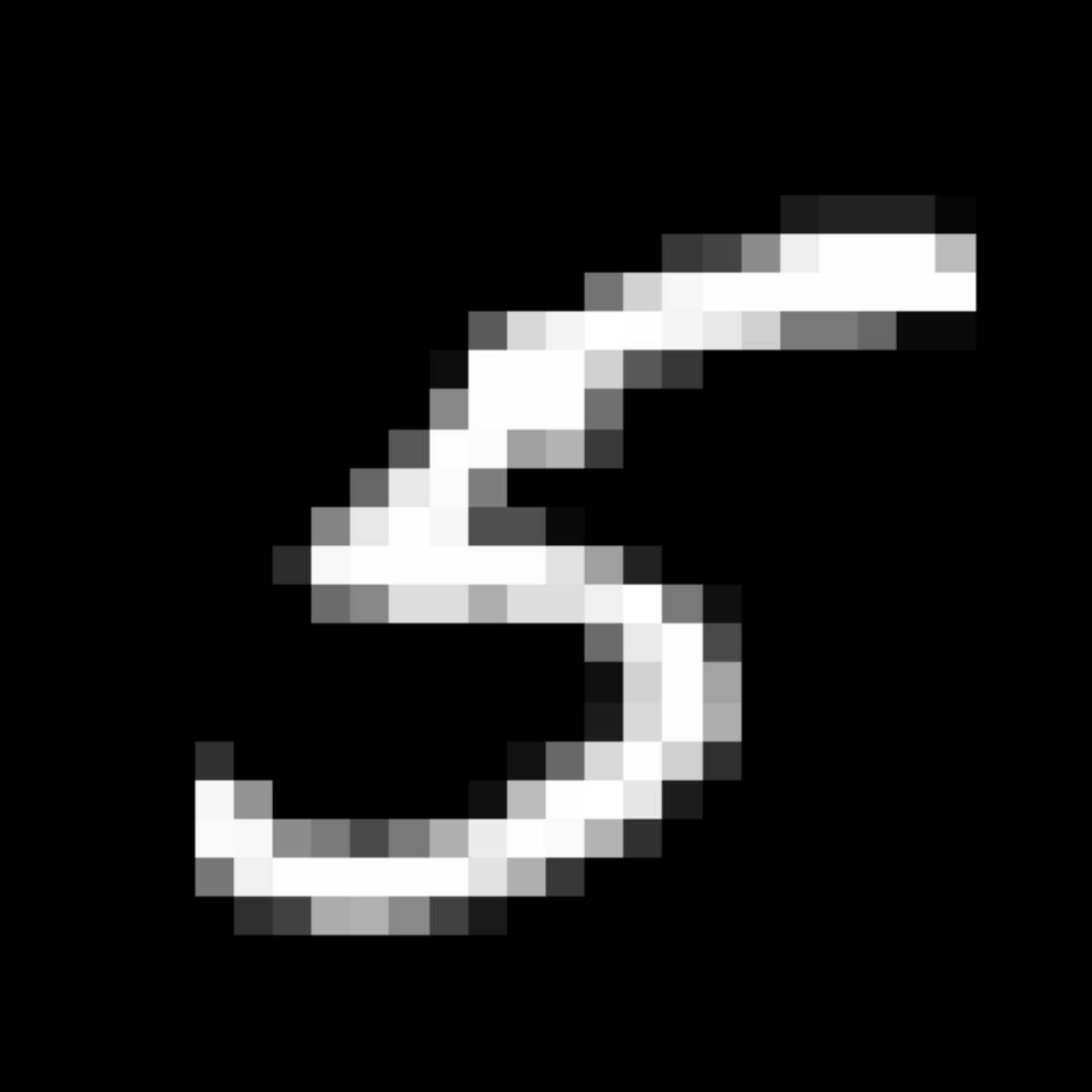}  & \mnist{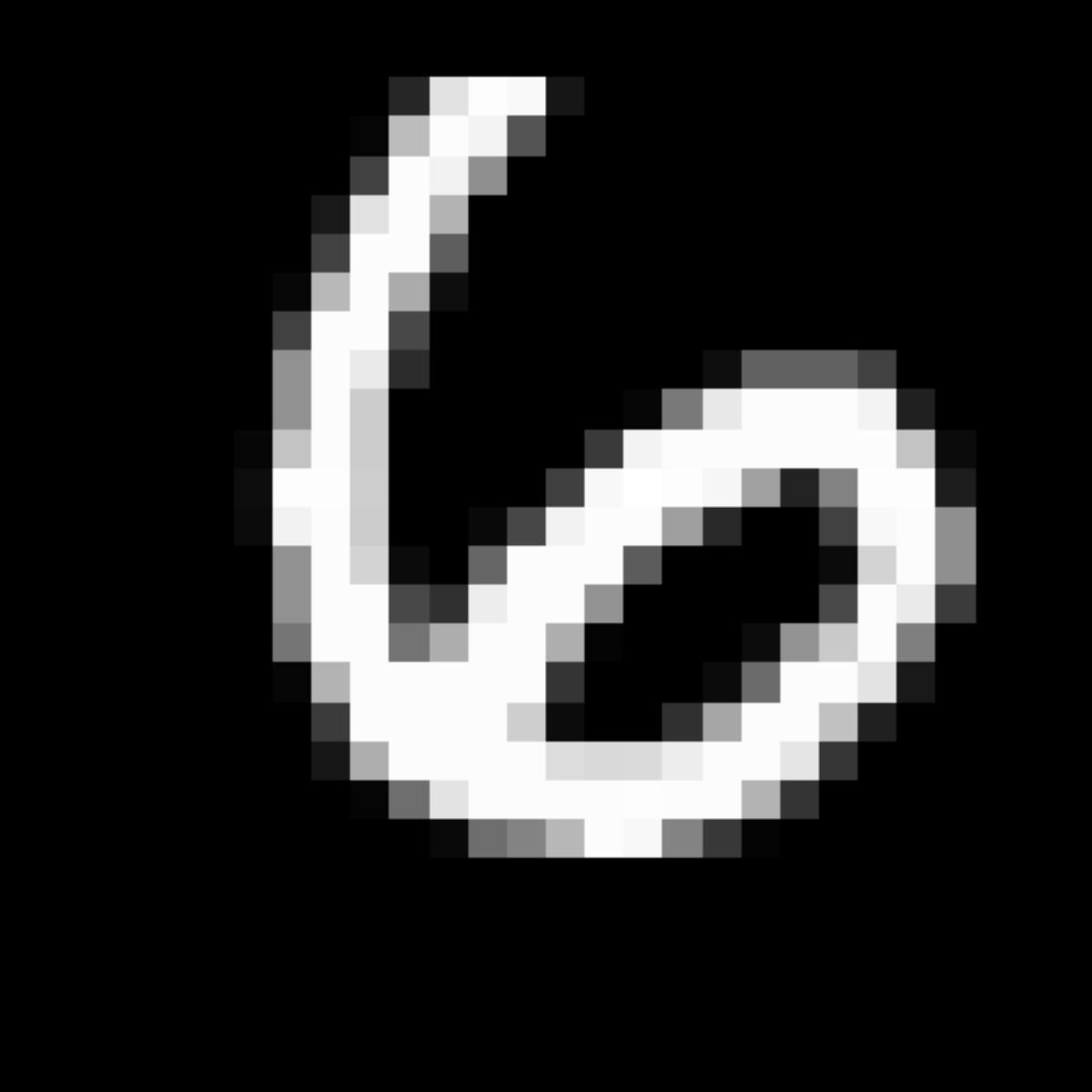}  & \mnist{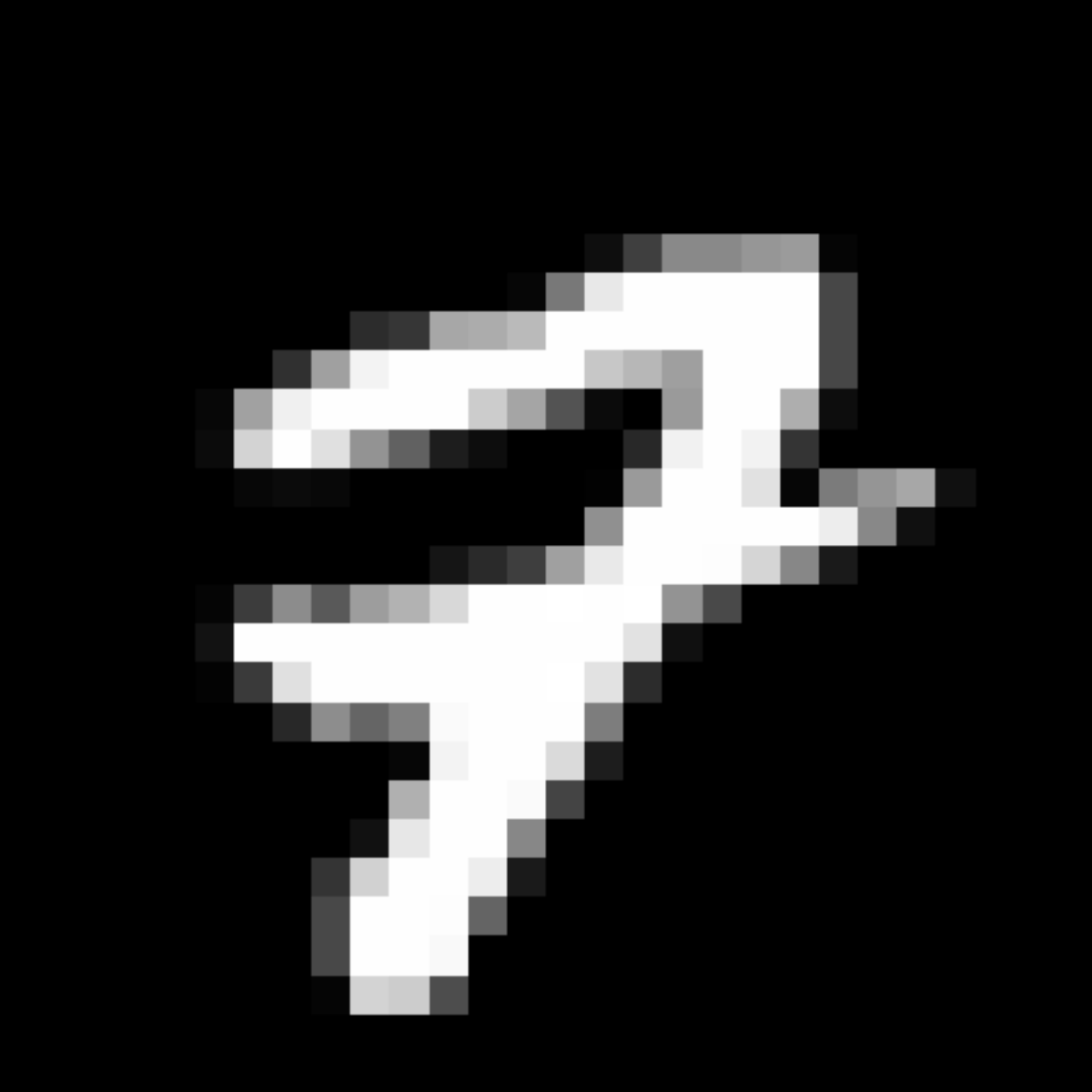}  & \mnist{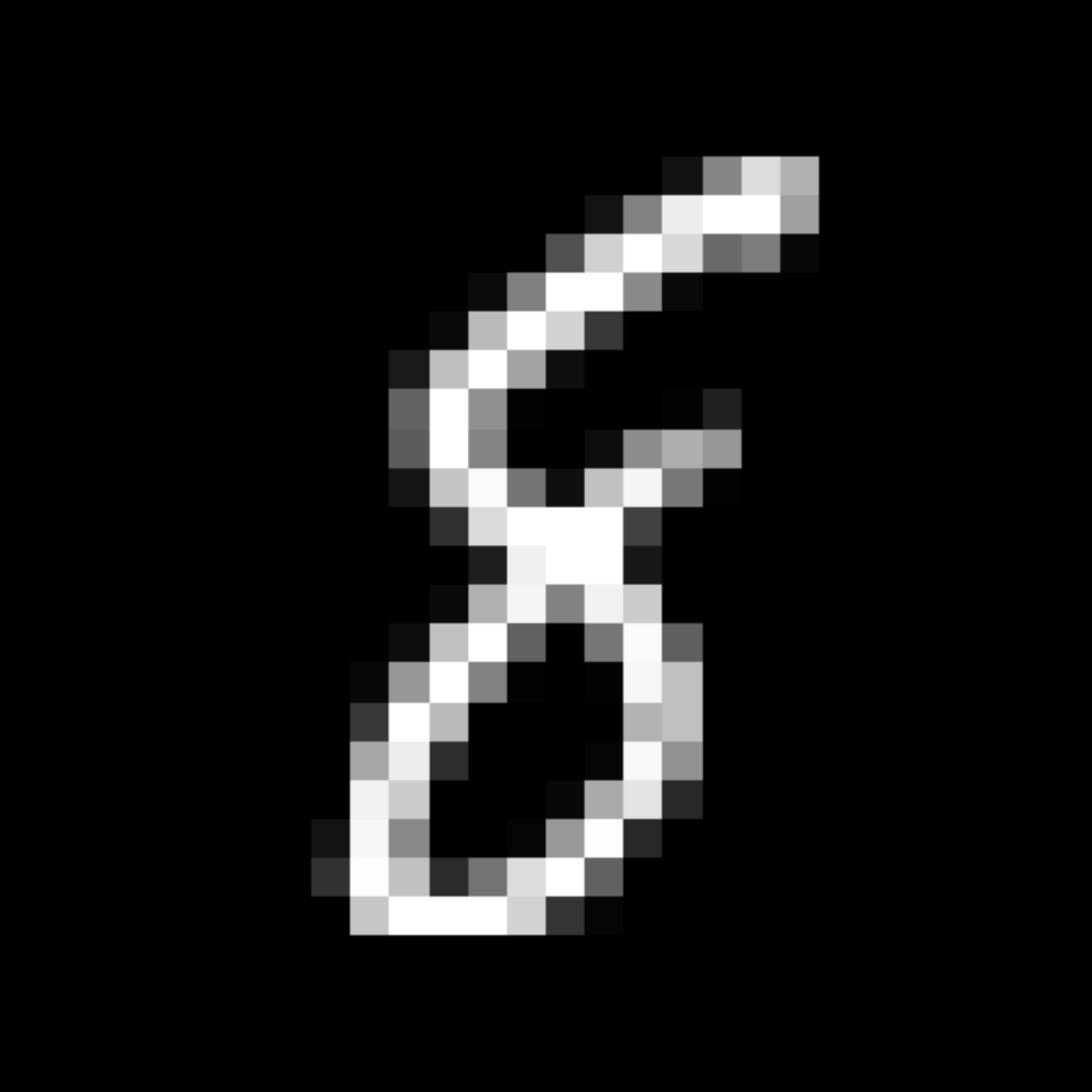}  & \mnist{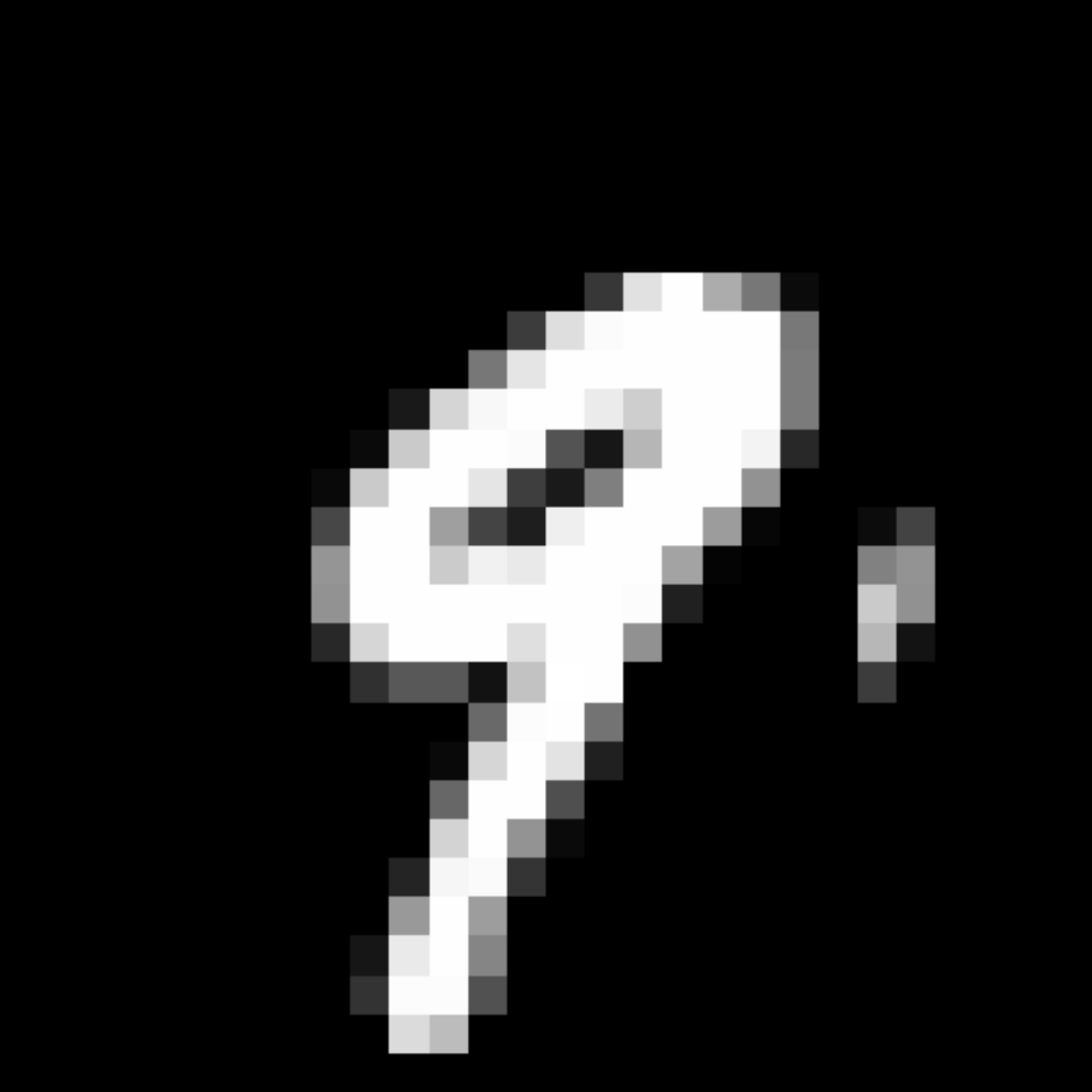}  \\ \hline
		FashionMNIST                            &  \cite{Xiao2017}   & $28$\,$\times$\,$28$ &          60\,000          &        10\,000        & \fmnist{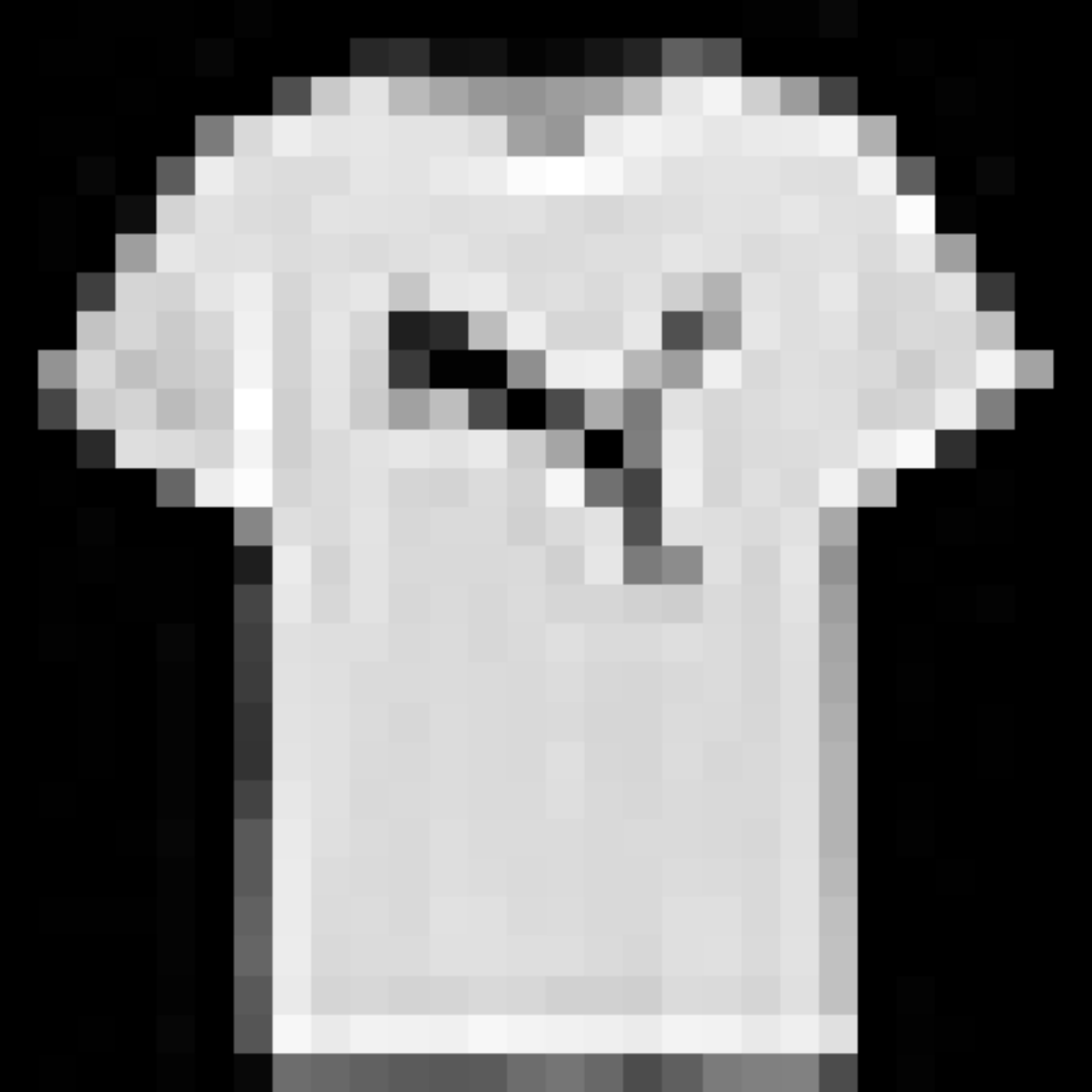} & \fmnist{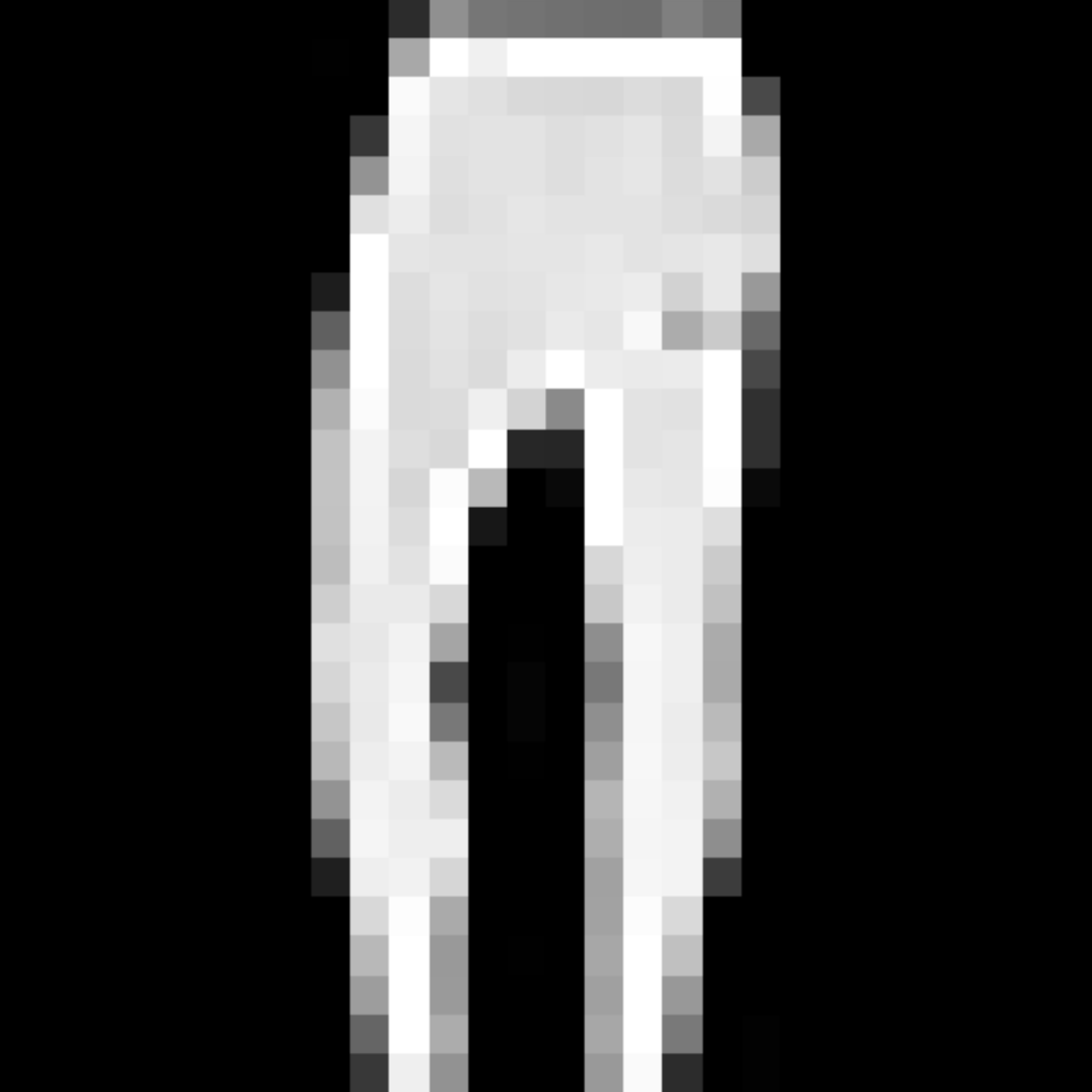} & \fmnist{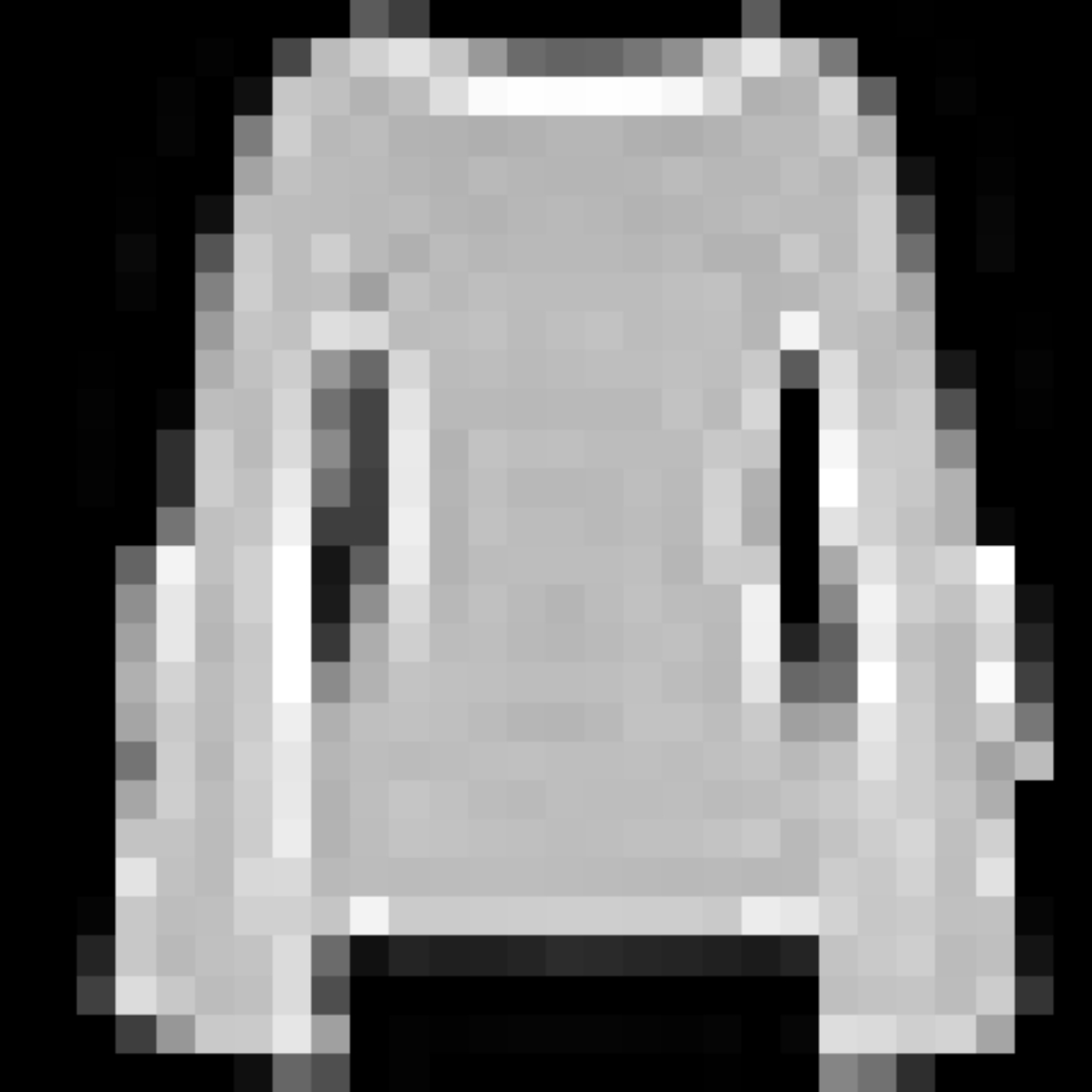} & \fmnist{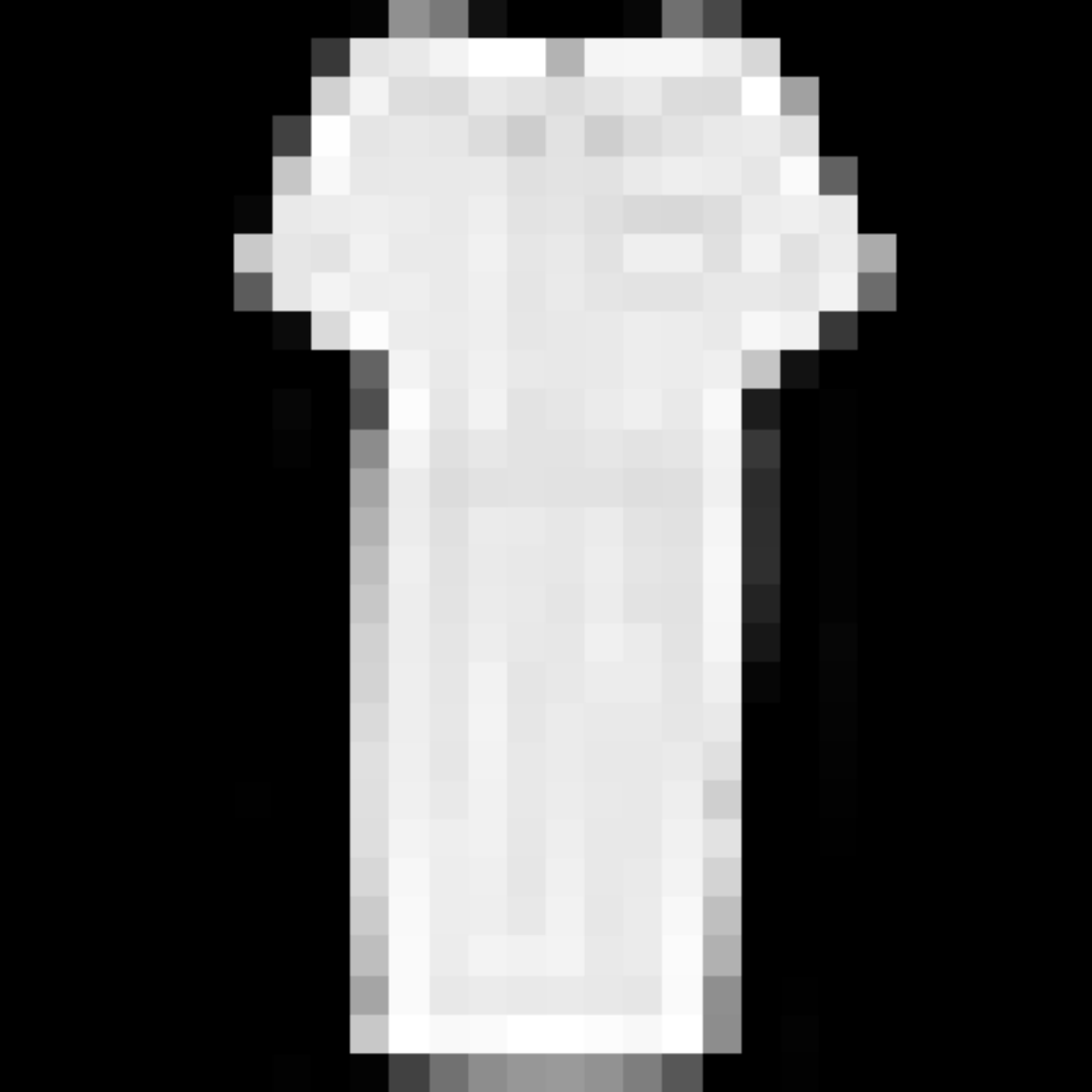} & \fmnist{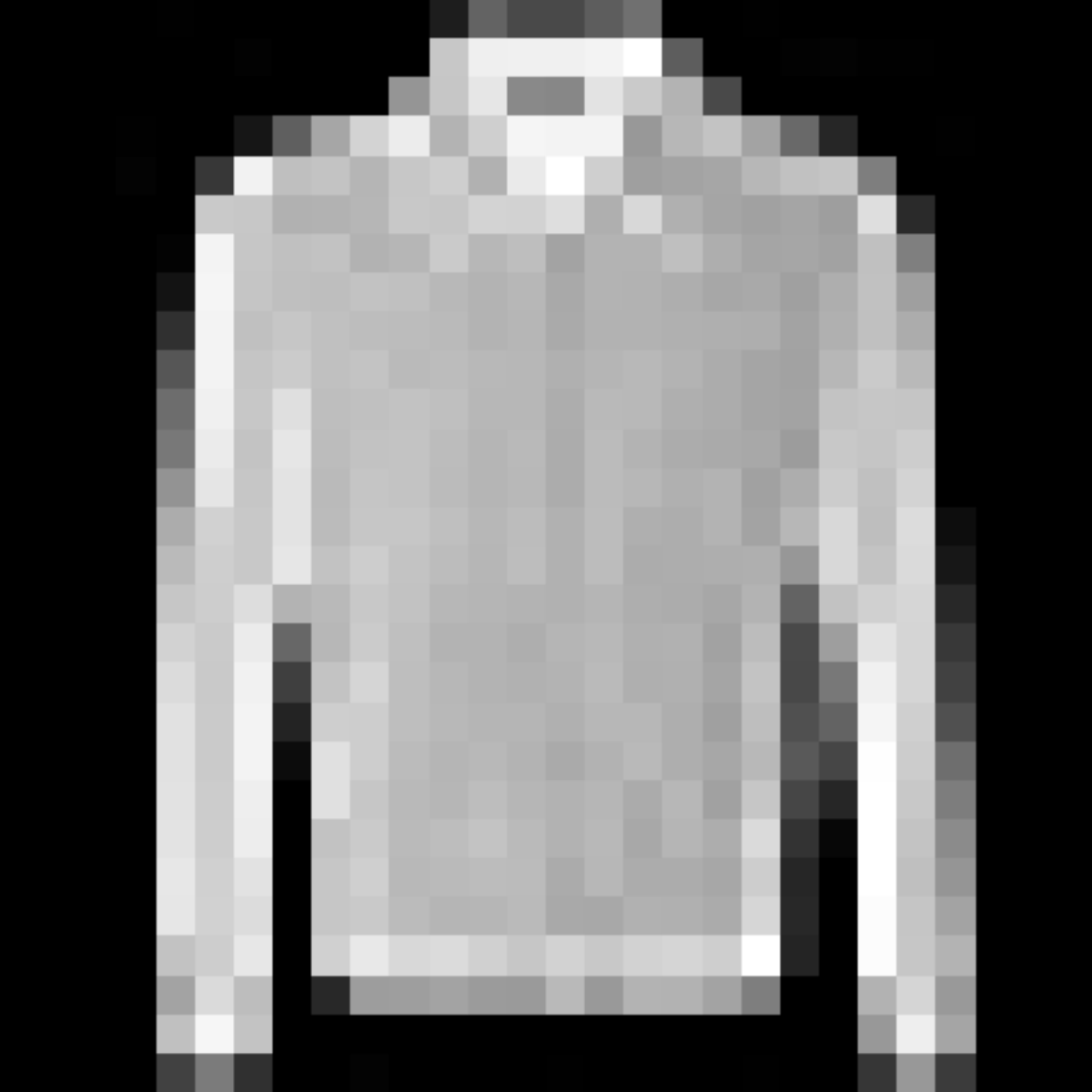} & \fmnist{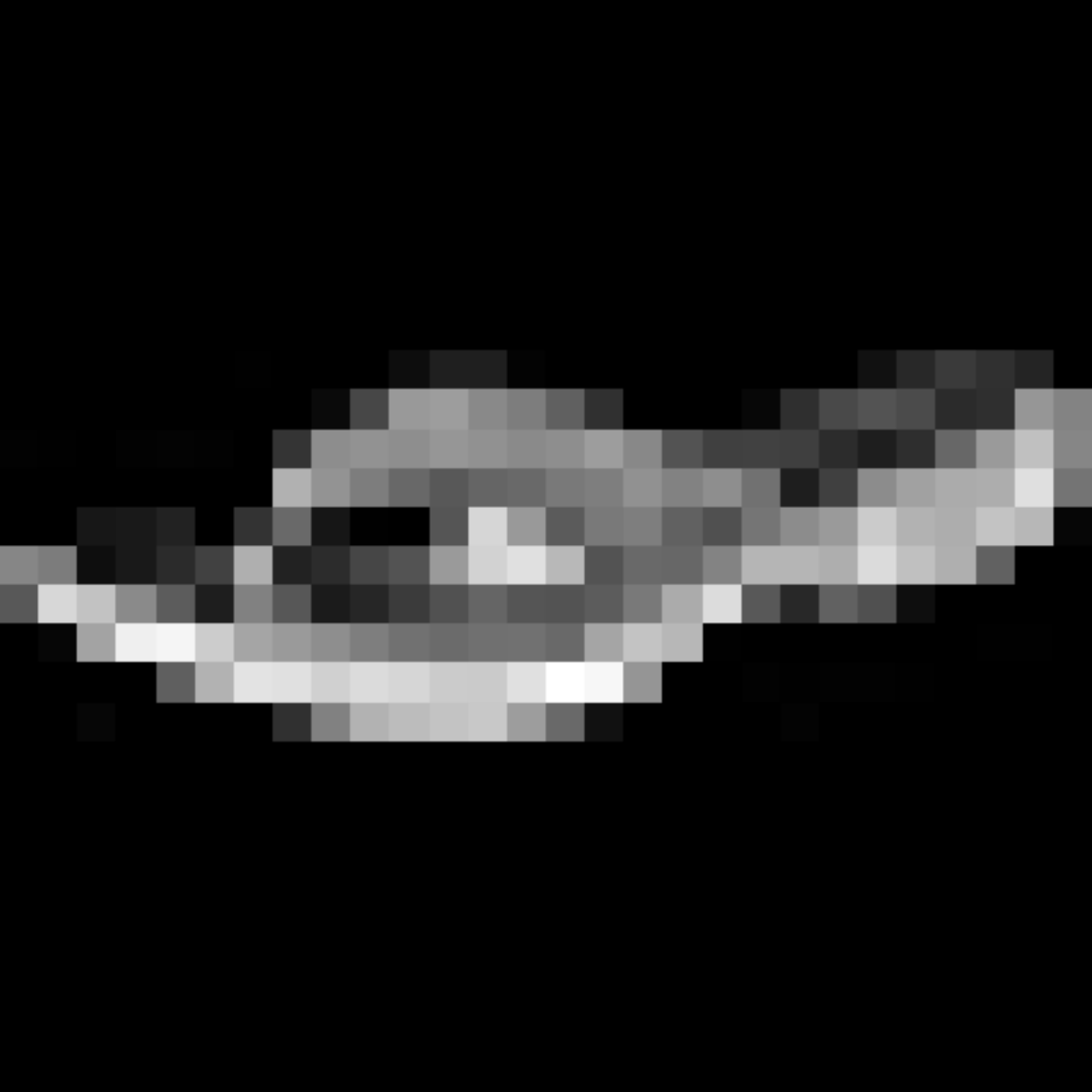} & \fmnist{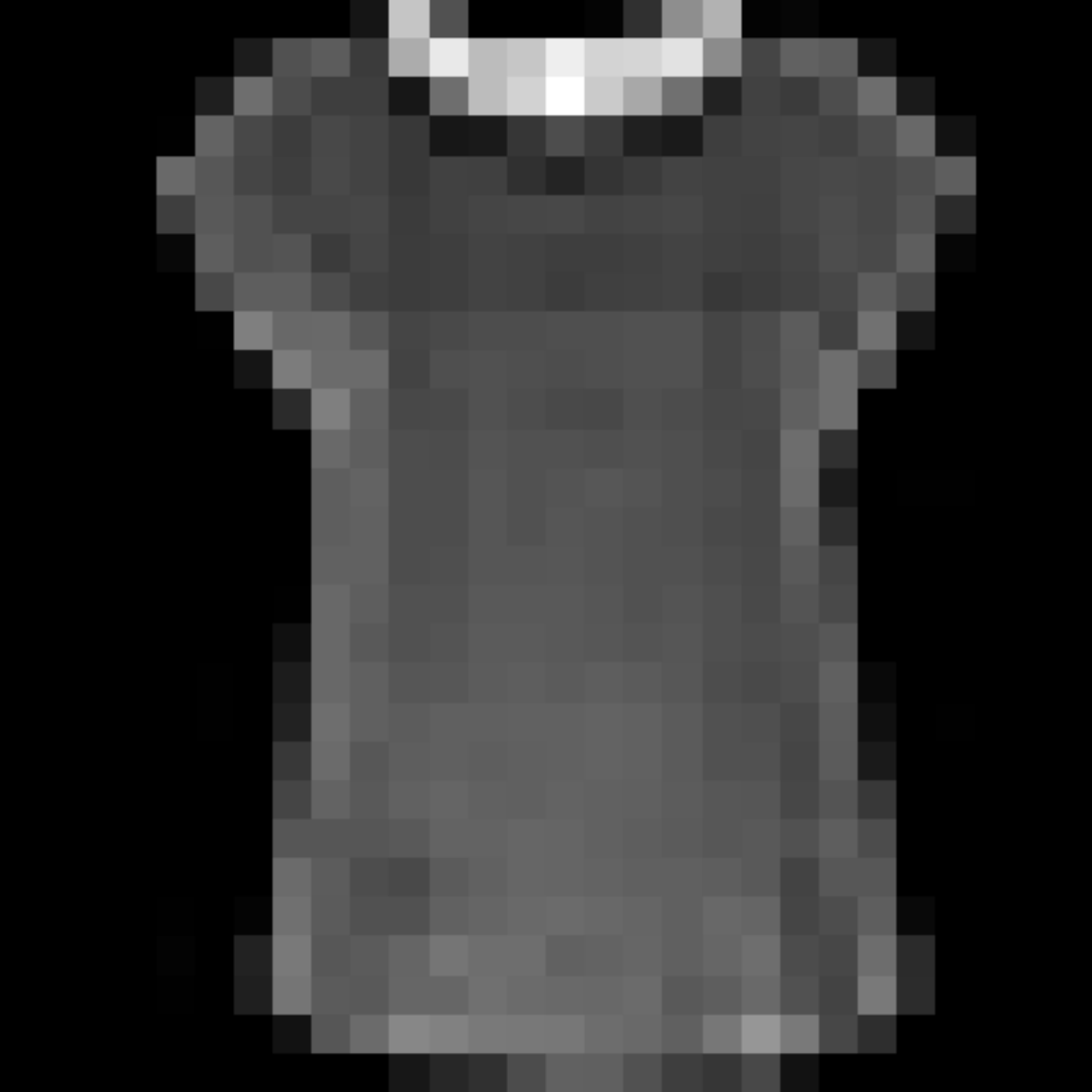} & \fmnist{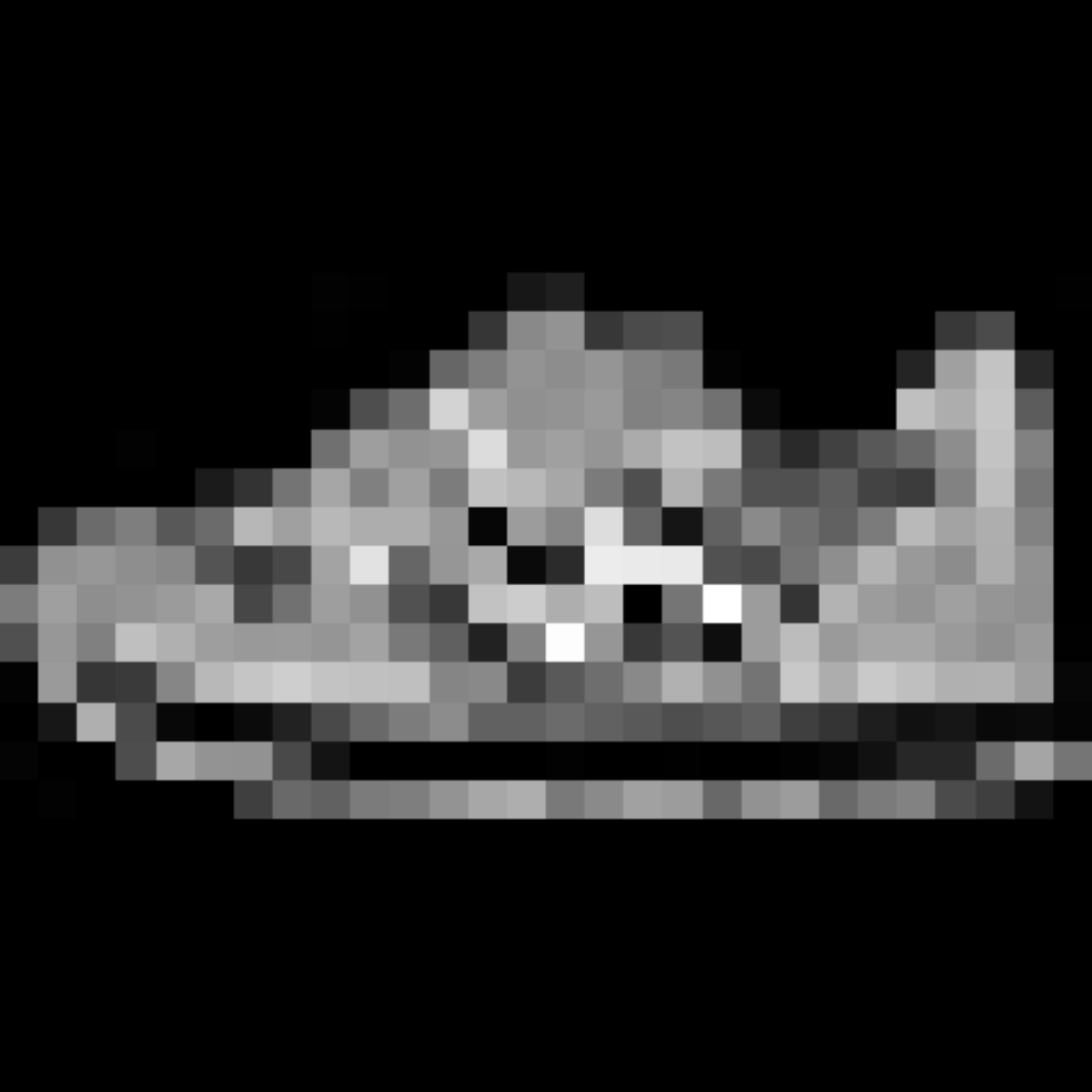} & \fmnist{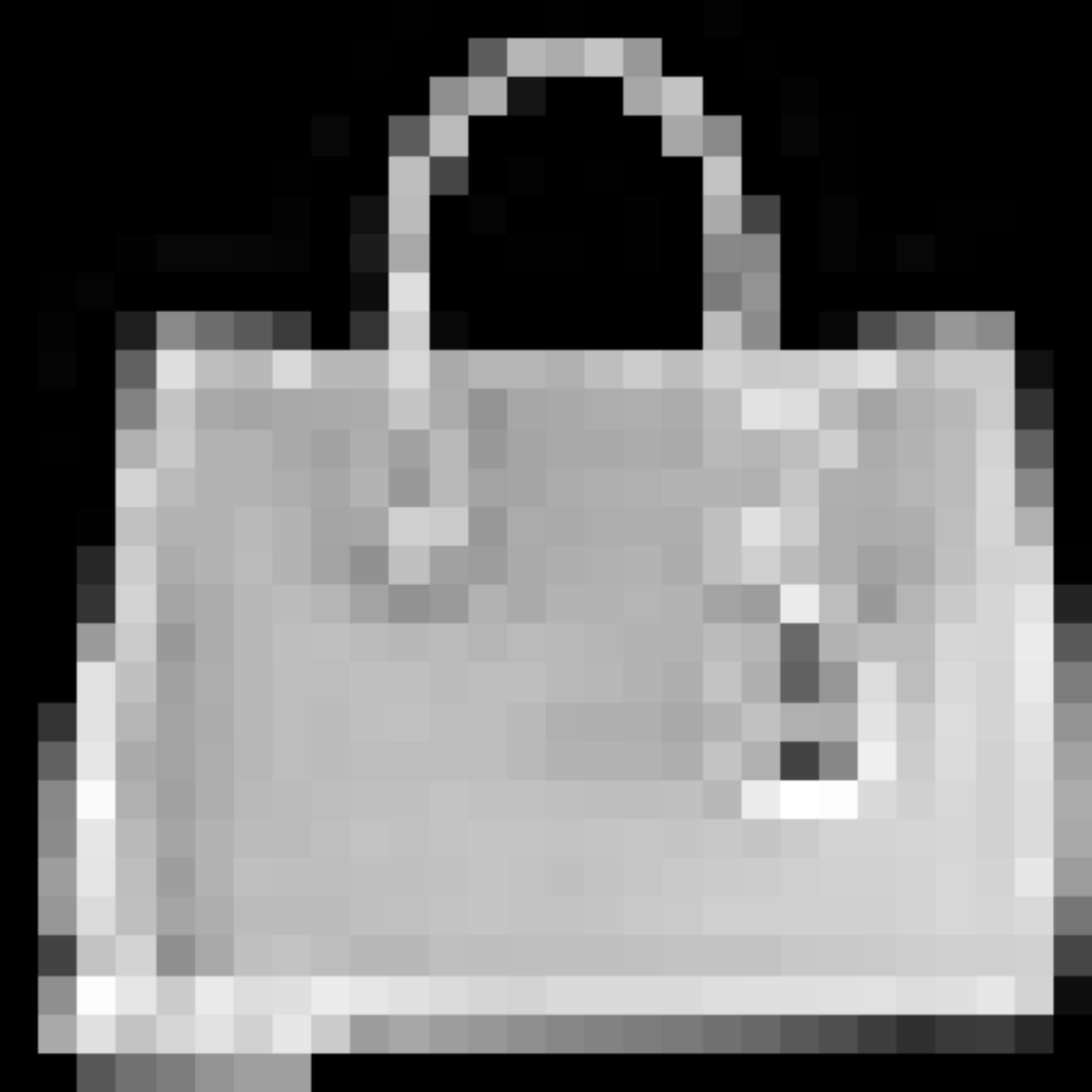} & \fmnist{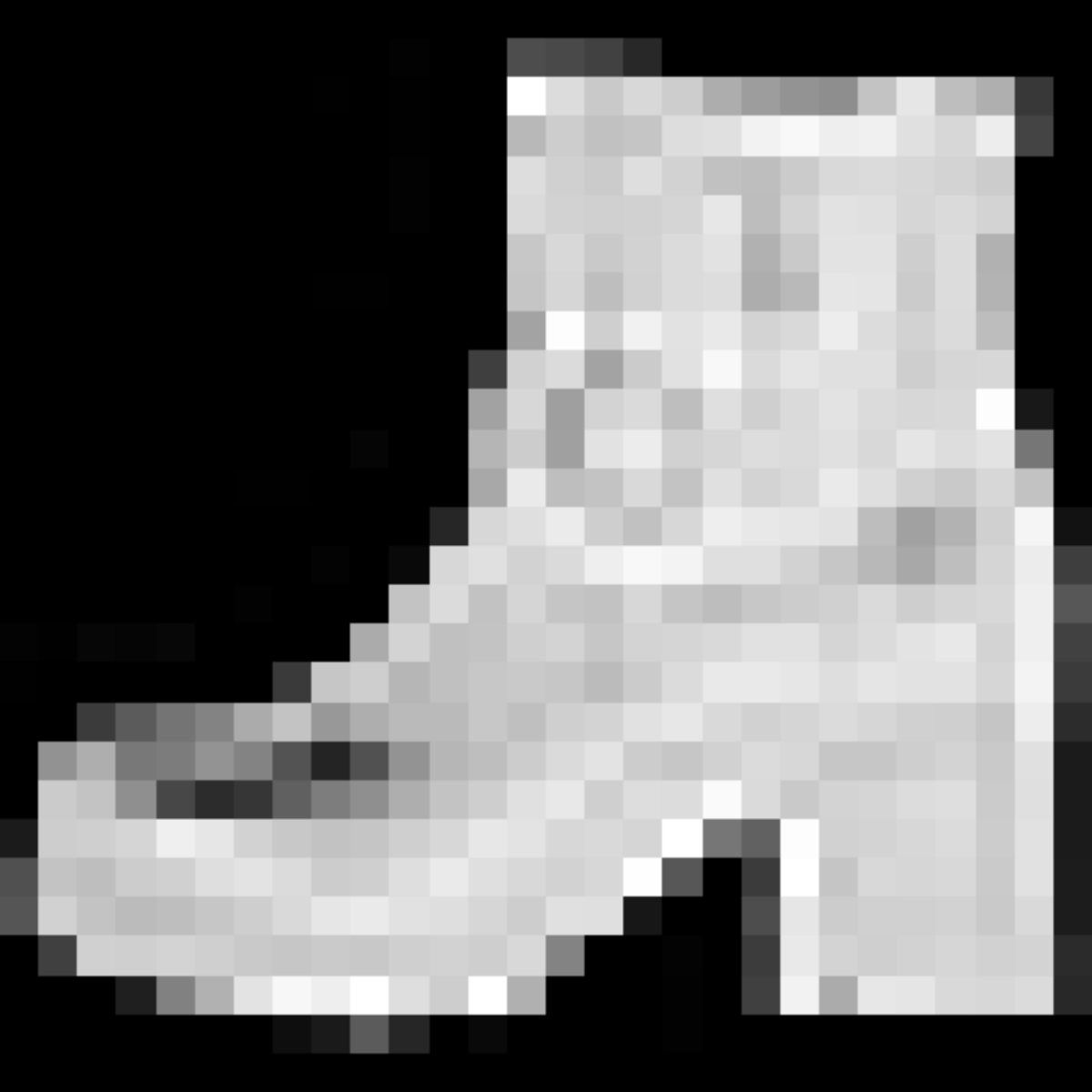} \\ \hline
		Devanagari                              & \cite{Acharya2015} & $32$\,$\times$\,$32$ &          18\,000          &   \phantom{1}2\,000   & \devan{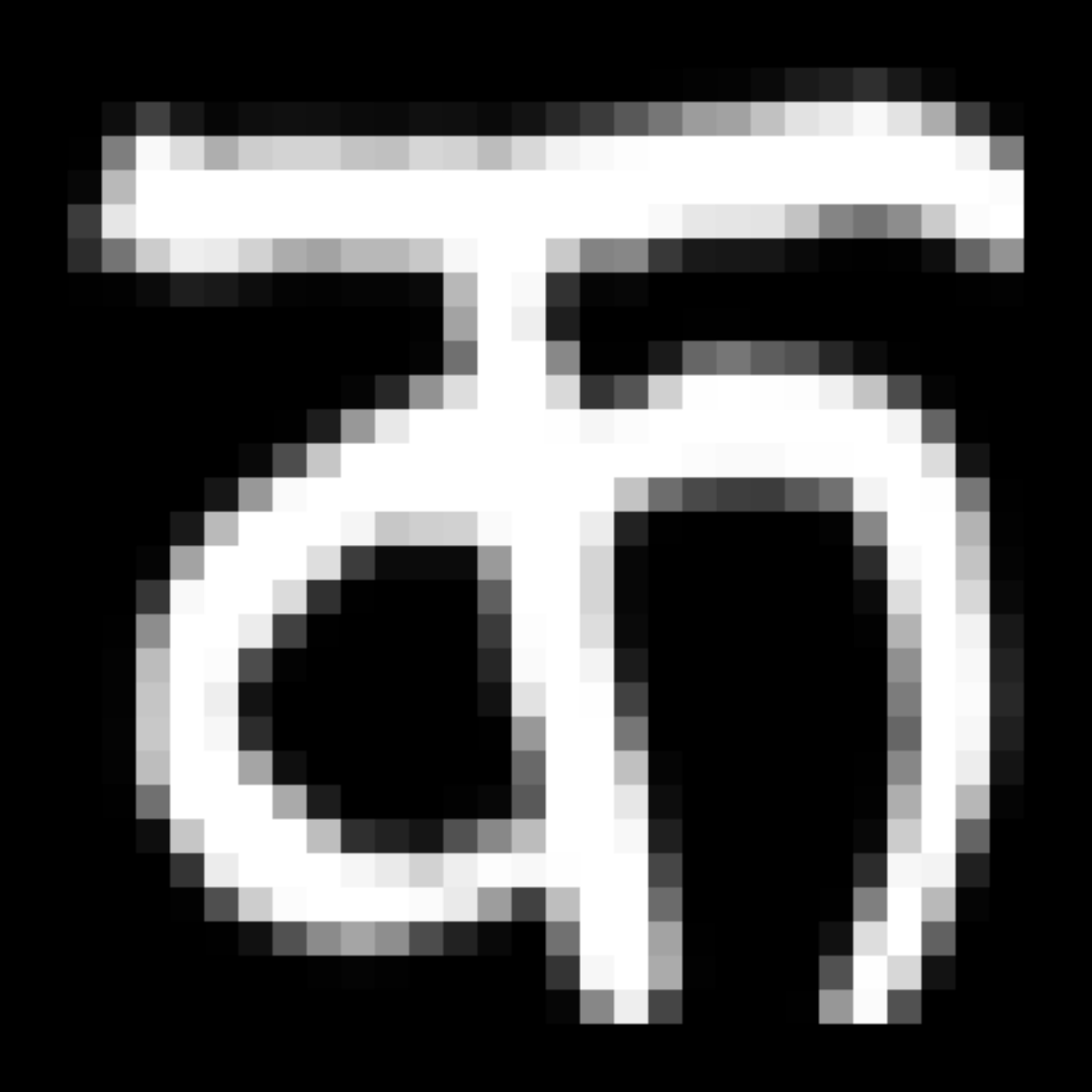}  & \devan{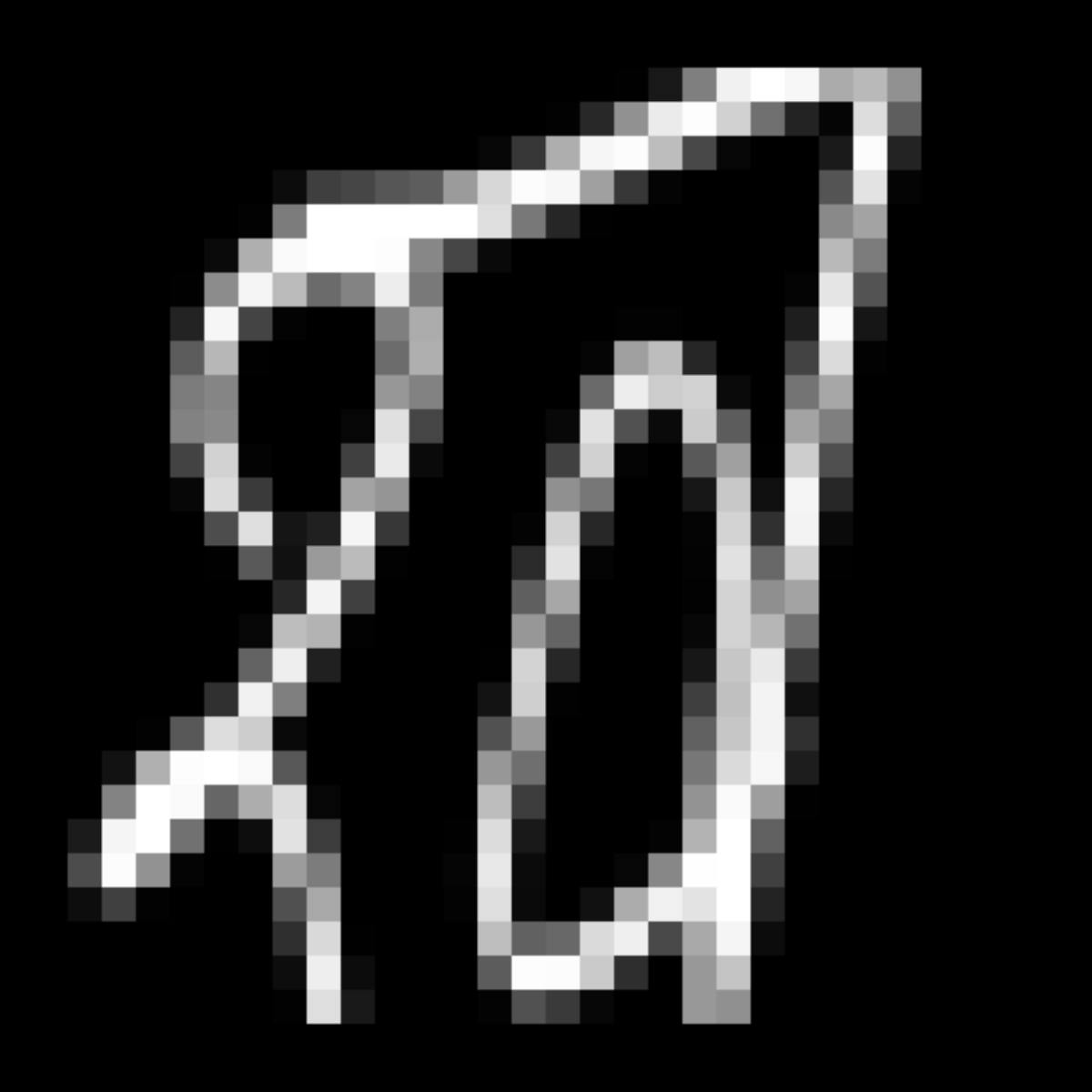}  & \devan{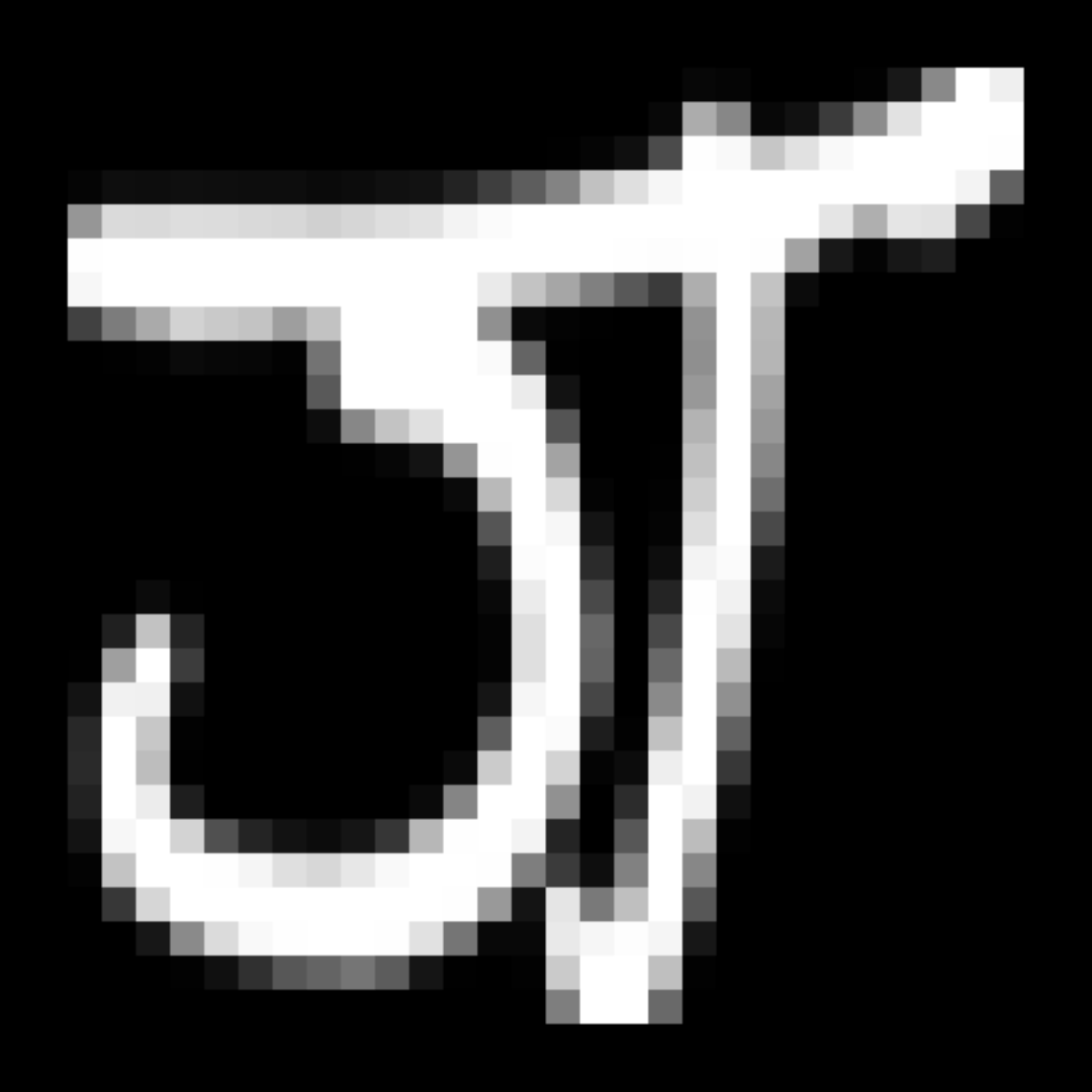}  & \devan{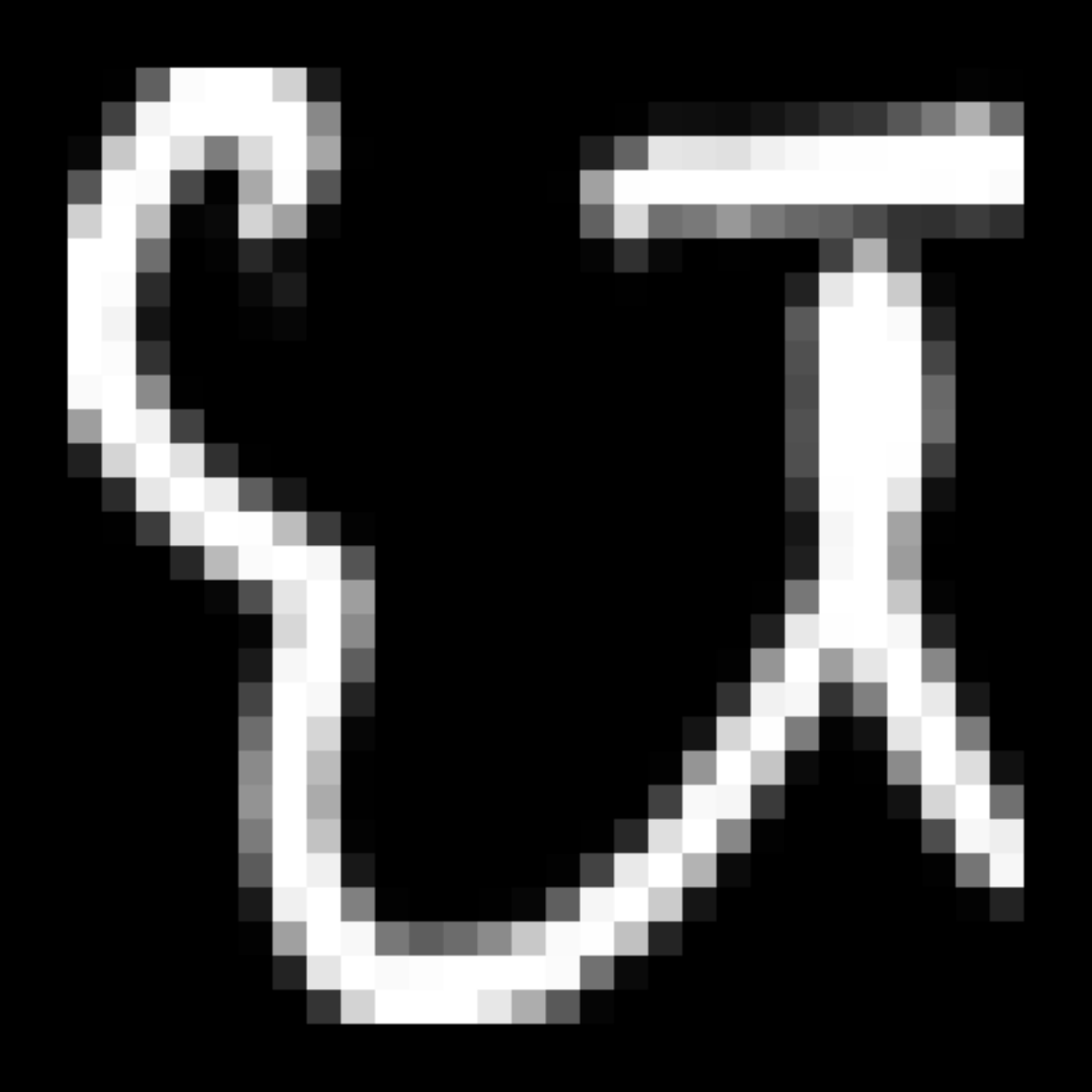}  & \devan{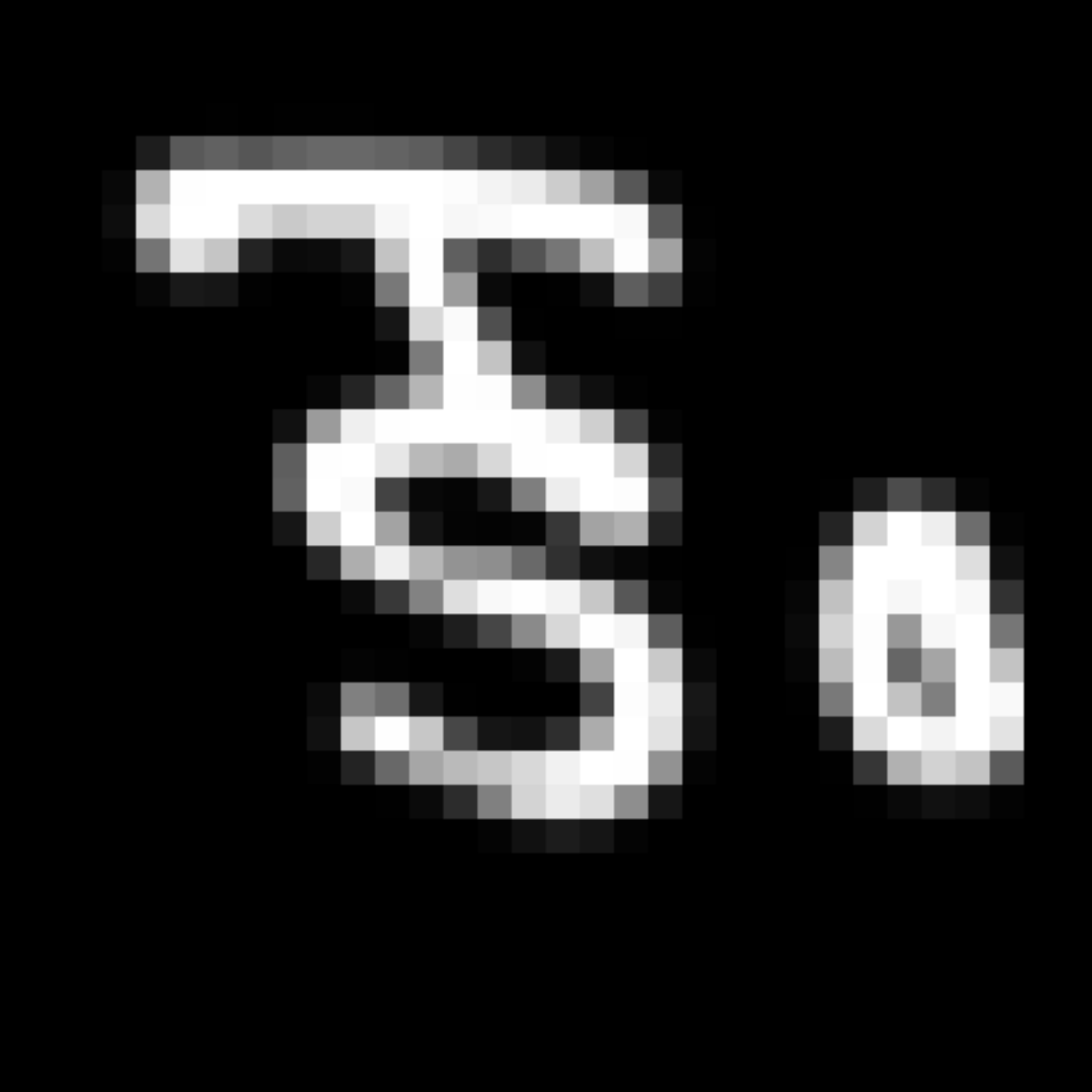}  & \devan{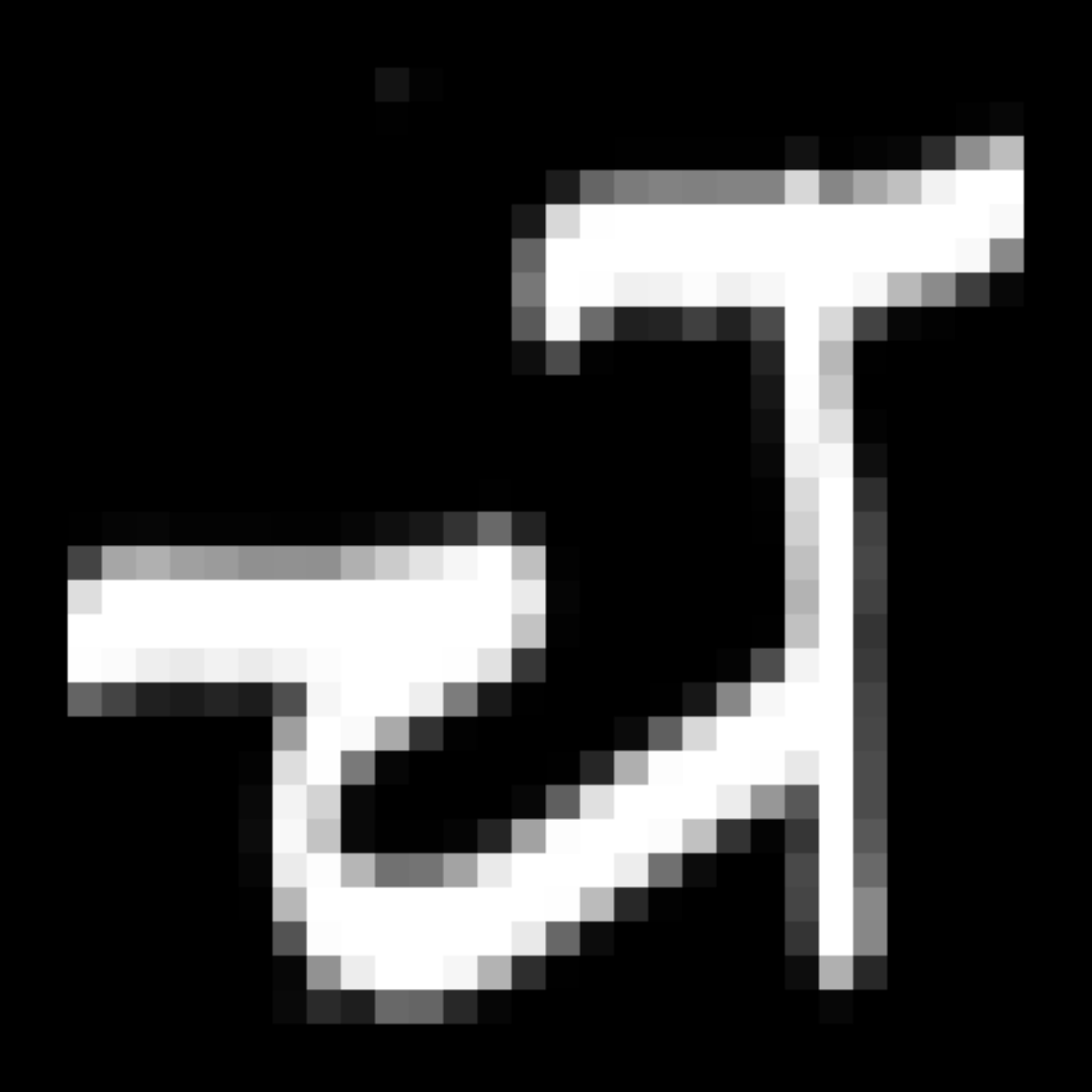}  & \devan{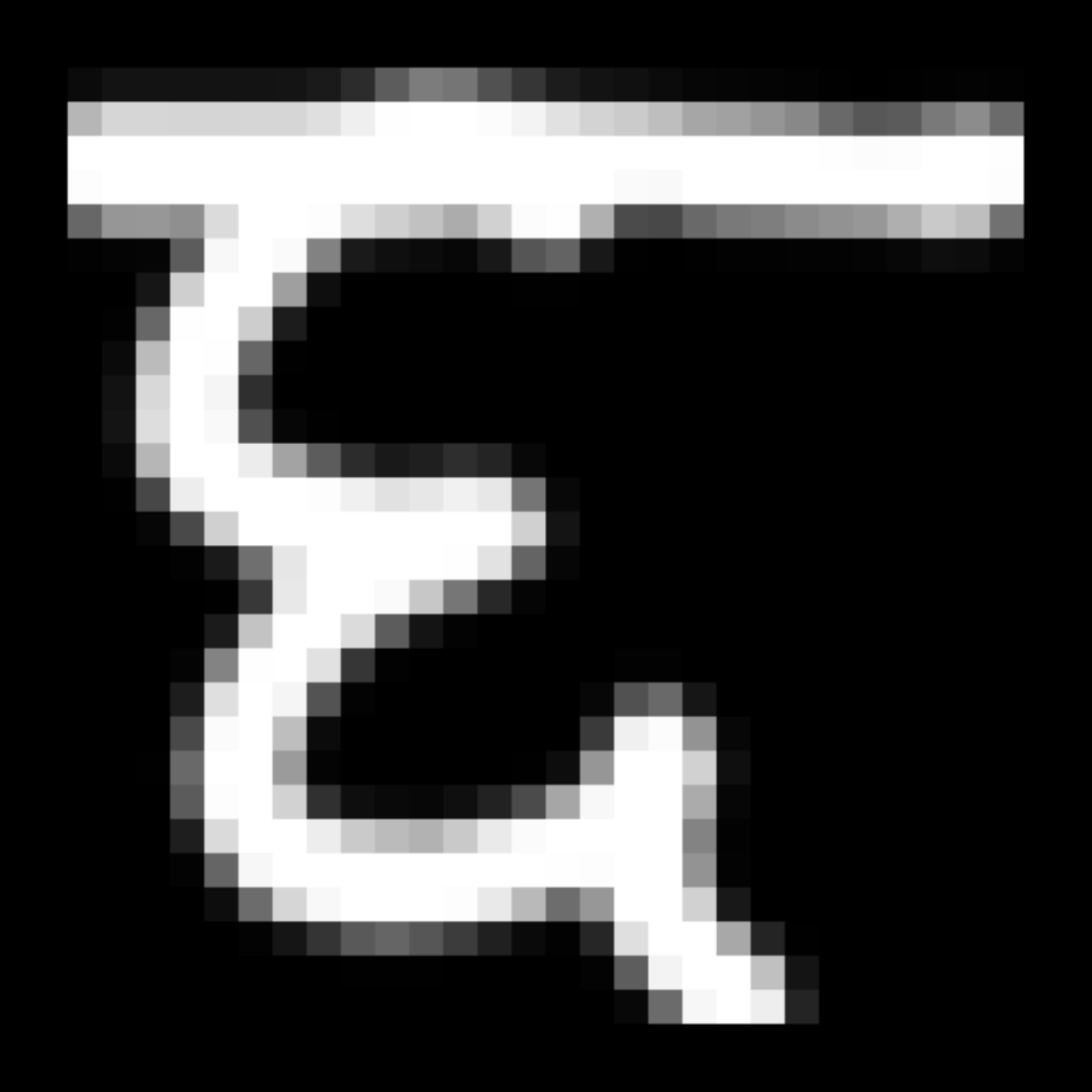}  & \devan{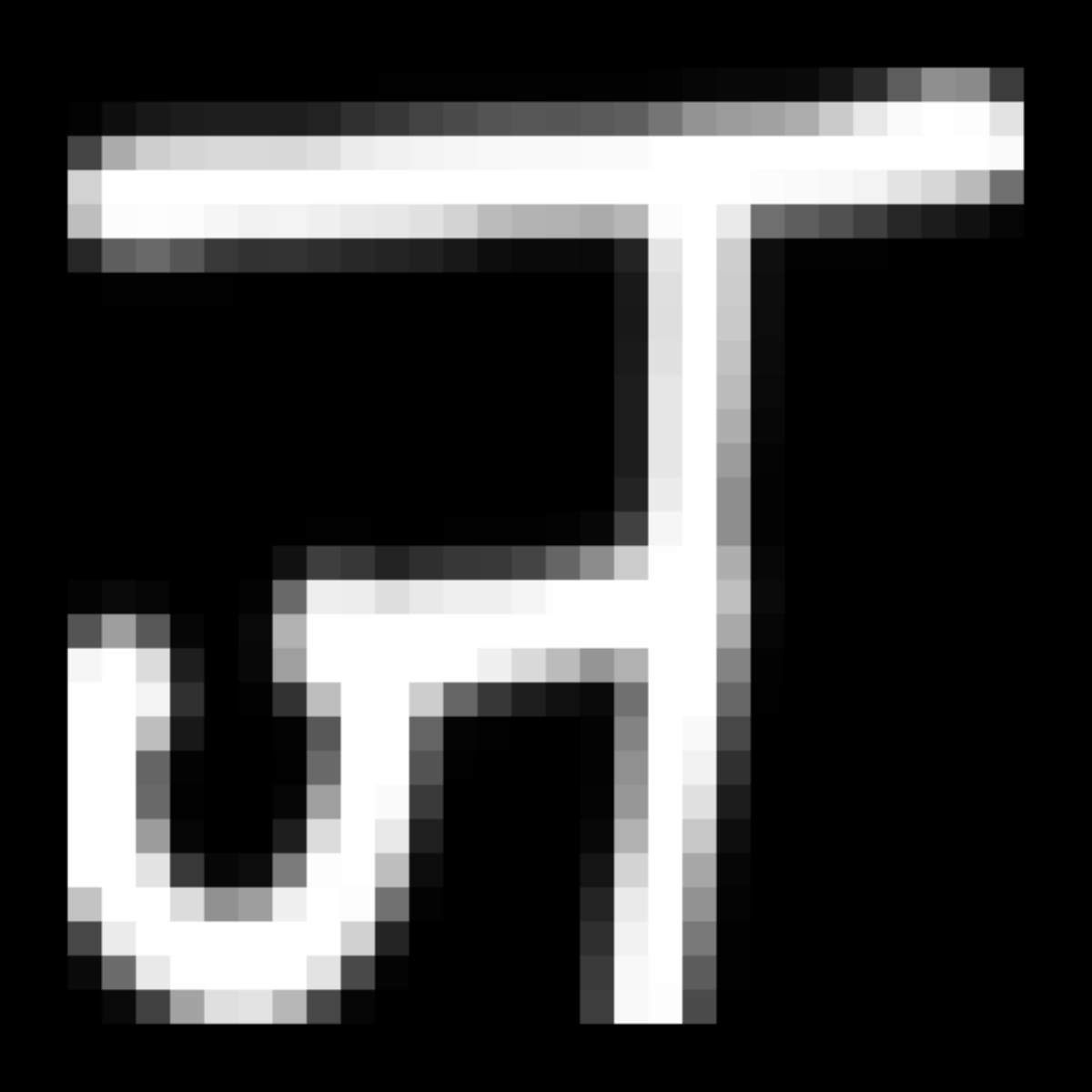}  & \devan{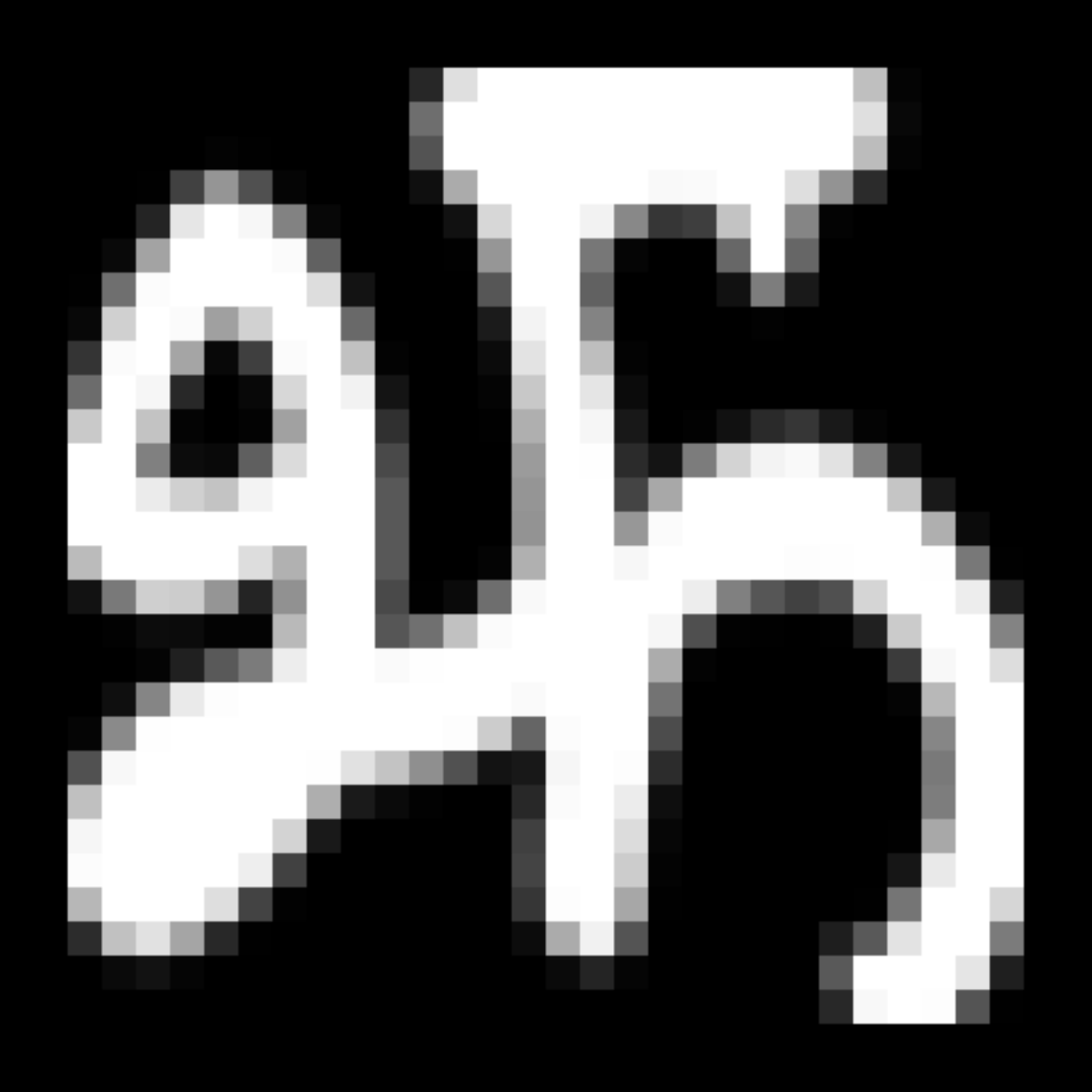}  & \devan{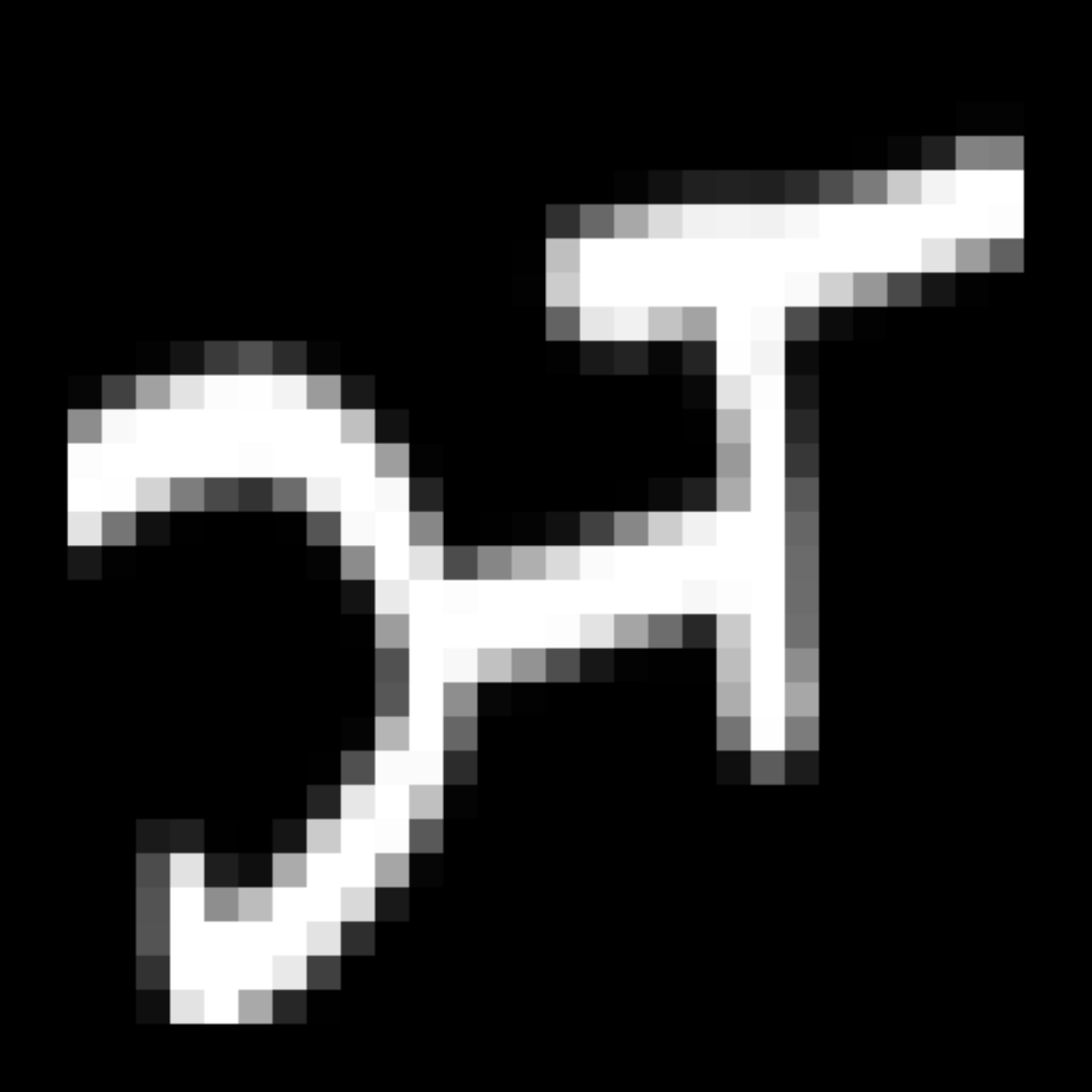}  \\ \hline
	\end{tabular}
\end{table*}
\section{The Gaussian Mixture Replay Model}\label{sec:GMR}
\noindent In this section, we provide details concerning the proposed GMR model. 
We put a particular emphasis on extensions/modifications of the GMM component to allow for a pseudo-rehearsal mechanism that performs well in practice.
For implementing and training the GMM as core component, we use the approach (and code, in a modified form) presented in \cite{Gepperth2019}.
This allows to train GMMs by Stochastic Gradient Descent (SGD) on high-dimensional data.
The latter is a hard requirement for the presented work, since all datasets are image-based and contain a significant number of samples.
The size of the used datasets makes the Expectation-Maximization (EM) approach to train GMMs unfeasible, which is why the efficient SGD training procedure proposed in \cite{Gepperth2019} is advantageous.
\par
GMR is a layered model containing a folding GMM and classification layer, see \cref{fig:gmr}. 
This is discussed in the following sections.
All layers expect their inputs to be four-dimensional tensors in NHWC format, as it is usual for CNNs.
\subsection{Folding Layer}\label{sec:fold_layer}
\noindent The folding layer performs a part of what a convolutional layer (in CNNs) would be doing, namely to extract organized slices from its input. 
Folding layers are parameterized by receptive field/filter sizes $f_X,\,f_Y$ and strides $\Delta_X, \Delta_Y$.
For an input tensor of the dimensions $N\!\times\!H\!\times\!W\!\times\!C$, a folding layer produces an output tensor of the dimensions $N\!\times\!H'\!\times\!W'\!\times\!C'$:
\begin{equation}
	\begin{split}
	H' & = 1+\frac{H-f_Y}{\Delta_Y} \\
	W' & = 1+\frac{W-f_X}{\Delta_X} \\
	C' & = f_X f_Y C
	\end{split}
\label{eqn:WHC}
\end{equation}
These quantities result from \enquote{dumping} the flattened content of a receptive field into a single channel dimension (a single \textit{column}) of the output.
In a CNN, this would be followed by a scalar product with the CNN filters, which is replaced here by the processing steps of the subsequent GMM layer.
\subsection{Gaussian Mixture Model Layer}\label{sec:gmm_layer}
\noindent GMM layers contain a single instance of a Gaussian Mixture Model. 
GMMs describe data samples $\vec{x}$ as a mixture of $K$ Gaussian densities $\mathcal N$, see \cref{eqn:density}.
\begin{equation}
	p(\vec x) = \sum_{k=1}^{K} \pi_k \mathcal N (\vec{x}; \vec{\mu_k},\Sigma_k)
\label{eqn:density}
\end{equation} 
For an input tensor of the dimensions $N\!\times\!H\!\times\!W\!\times\!C$, this GMM instance processes all $HW$ columns of its input tensor independently. 
Thus, the GMM instance processes input vectors $\vec{x}$ of dimension $C$.
The posterior probability (or responsibility) of the GMM, which we treat as the GMM layer's output to the classifier layer, is computed as: 
\begin{equation}
	\gamma_{j}(\vec{x}) = \frac{\pi_j \mathcal{N}(\vec{x};\vec{\mu_j},\Sigma_j)}{\sum_k \pi_k \mathcal{N}(\vec{x};\vec{\mu_k},\Sigma_k)}.
\label{eqn:resp}
\end{equation} 
Thus, responsibilities have the dimensions of $N\!\times\!H\!\times\!W\!\times\!K$. 
They are normalized and bounded in the interval $[0,1]$.
\par
To learn the distribution of inputs from data, we use the SGD procedure from ~\cite{Gepperth2019}.
This is, however, a technical detail, and trained GMMs behave like any other GMM trained by Expectation-Maximization (EM).
In particular, \cref{eqn:density} can be used for detecting outliers whose (log-)probability is low. 
For sampling, a value is drawn from a component distribution $\mathcal{N}(\vec{x}; \vec{\mu_k},\Sigma_k)$. 
$k$ is drawn from a multinomial distribution parameterized by the weights $\vec{\pi}$ of the GMM.
In GMR, the downstream linear classifier layer can provide a control signal to the GMM layer $\vec t$ for sampling.
In this case, the parameter to the multinomial distribution is $\vec t$, and sampling is performed for each column independently, resulting in a sampling result of the same dimension as the input.
\subsection{Linear Classification Layer}\label{sec:lin_layer}
\noindent Since GMMs are inherently trained in an unsupervised manner, they cannot compute class labels. 
For the assignment of labels to GMM outputs, we use a linear classifier layer that performs the operation $\vec{\gamma}W\!+\!\vec{b}$ on the responsibilities $\vec{\gamma}$.
The softmax function, which is usually applied beforehand, is already included in the responsibilities, see \cref{eqn:resp}.
For training, either cross-entropy or mean squared error (MSE) can be used as loss function, in combination with SGD.
For generating a control signal to the GMM layer, a one-hot vector $\vec{o}$ representing the desired class is created and transformed as $\vec{t}=\text{norm}_{[0,1]}(\vec{o}-\vec{b})W^\top$.
Thus, it is possible to select components from the overlying GMM layer to enable conditional sampling.
\subsection{Classification Process Example}\label{sec:classification}
\begin{figure}[htb!]
	\centering
	\includegraphics[width=\linewidth]{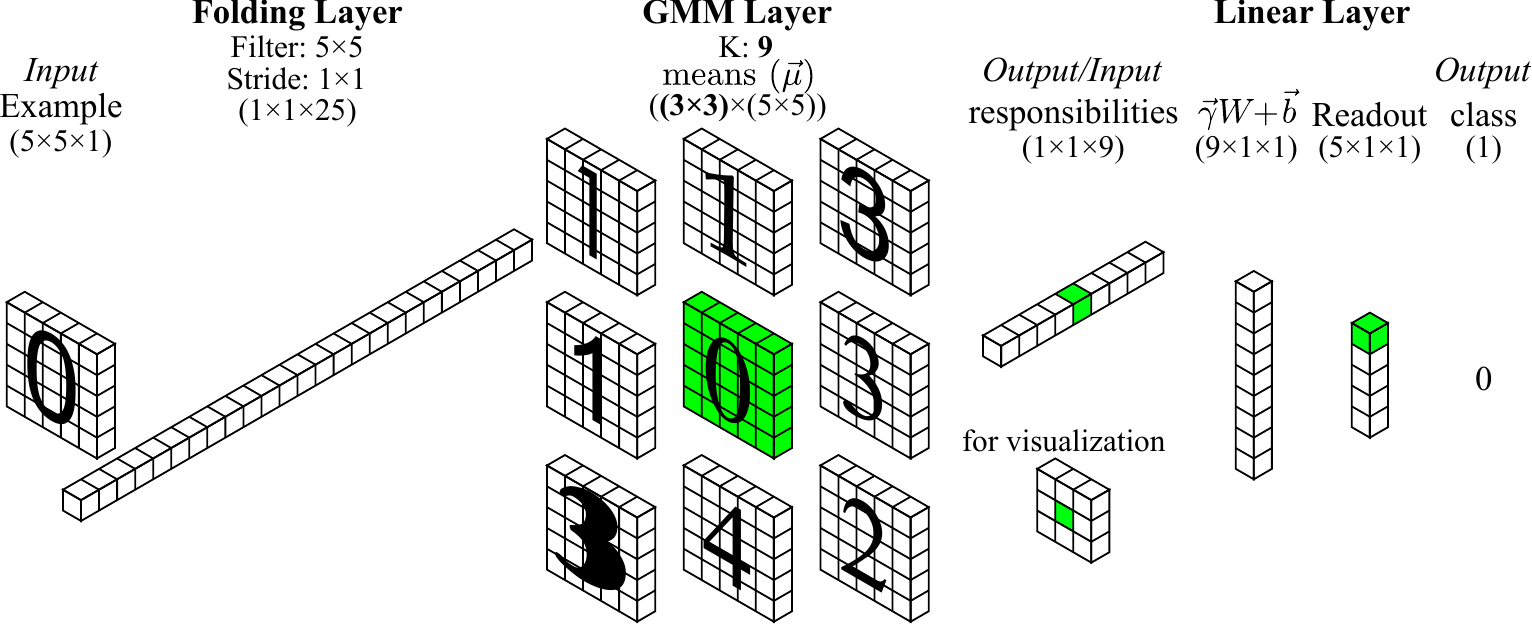}
	\caption{Visualization of the classification process of a GMR model.}
	\label{fig:train}
\end{figure}

\noindent In the following section, the classification process by the GMR is described and also visualized as an example (see \cref{fig:train}). 
Let us assume that the input tensor has the following format: $N$\,$=$\,$1$, $H$\,$=$\,$5$, $W$\,$=$\,$5$ and $C$\,$=$\,$1$.
For the sake of simplicity, we omit the dimension $N$ of shape $1$. 
We are left with a single $5$\,$\times$\,$5$\,$\times$\,$1$ gray-scale image as input $\vec{x}$ (see \cref{fig:train} left).
This leads to a given GMR which consists of a folding layer, a GMM layer and a classification layer (central part).
\par
The folding layer unfolds the input signal with a receptive field of the size $f_X$\,$=$\,$f_Y$\,$=$\,$5$. 
It follows that striding is not used.
This shows that the image is transformed line by line into a $1$D vector, which is passed to the GMM layer.
\par
The visualized GMM layer consists of $K$\,$=$\,$9$ components arranged in a $3$\,$\times$\,$3$ grid.
Each component is composed of $\vec{\mu_k}$ (the mean) and $\Sigma_k$ (the variances) with the shape of the input signal ($1$\,$\times$\,$25$).
For simplicity, only the means are displayed, which in turn were reshaped from a $1$D vector (same shape an input signal) into a $2$D image.
We assume that the GMR has already converged and derived the following data distribution from the training data.
Note that the prototypes are presented in an exaggerated way.
The central component (green highlighted) is the one that has the largest association for the input signal (0 sample), which is why the responsibility ($\gamma_k$) for this prototype is the highest (also green). 
For clarity, all responsibilities ($\vec{\gamma}$) are additionally transformed ($3$\,$\times$\,$3$) to match the visualization of the GMM components.
\par
The linear layer receives the responsibilities ($\vec{\gamma}$) of the GMM layer.
Here, the class label is assigned based on the learned distribution.
The output of this layer, and thus, the GMR, is a class label associated with the input. 
\subsection{Sample Generation Example}\label{sec:sampling}
\begin{figure}[htb!]
	\centering
	\includegraphics[width=\linewidth]{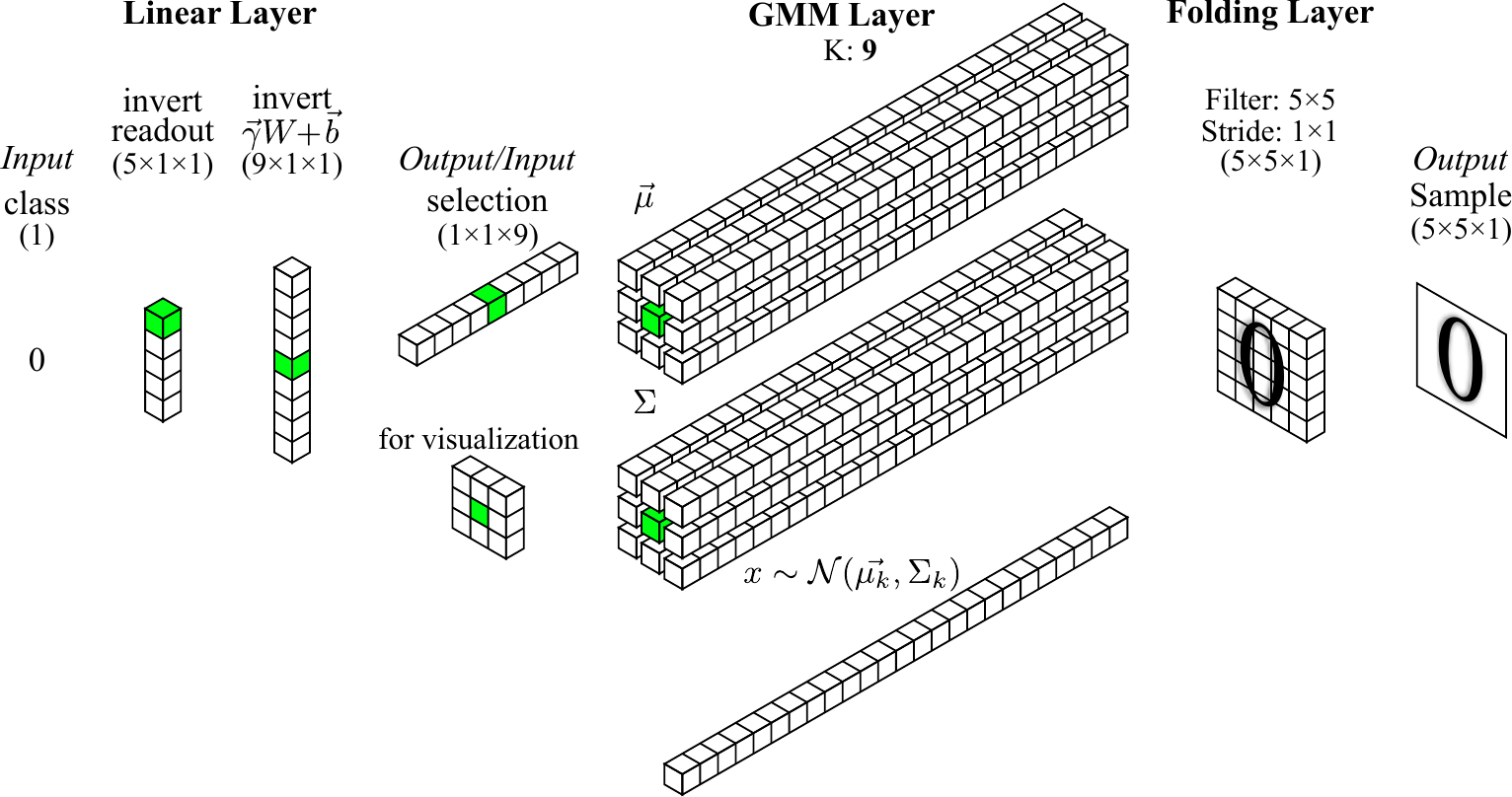}
	\caption{Visualization of the sampling process of a GMR model.}
	\label{fig:sampling}
\end{figure}

\noindent The inverse function for classification is the sample generation.
As described in \cref{sec:classification}, we assume that the model is already trained.
Again, we set $N$\,$=$\,$1$ and neglect this dimension for visualization.
In \cref{fig:sampling} the reverse data flow through the GMR is depicted.
To generate samples conditionally, the class from which a sample should be generated is used as input.
It is converted as one-hot vector (see \cref{fig:sampling} left).
The class from which an example is generated is highlighted in green.
\par
The linear layer is used to inverse the mapping of a GMM component to a class by generating a signal which is passed on to the GMM layer above.
For this, the inverse transformation is taken as described in \cref{sec:lin_layer}.
It is also highlighted in green and arranged for visualization ($3$\,$\times$\,$3$) as the components of the upper GMM layer.
\par
This time, for each GMM component the means ($\vec{\mu}$) and corresponding variances ($\Sigma$) in their true form are given.
We again assume that the GMR has already learned the distribution from the data (see \cref{fig:train} for an impression of the prototypes). 
The selection strategy also includes the choice of different components that simultaneously represent the same class.
This is achieved on the basis of a probability distribution depending on the signal of the classifier layer.
In this example, only one sample should be created and only the central component (green highlighted) is responsible, which is why it is very likely to be selected for sampling.
A sample $x$ is drawn from the selected GMM component by sampling from a multivariate normal distribution $\mathcal{N}(\mu_k,\Sigma_k)$.
\par
The last step is to restore the original form of the input data (images).
The folding layer inverts the flattening process by reshaping  $x$ into the origin input shape ($5$\,$\times$\,$5$\,$\times$\,$1$).
\section{Elastic Weight Consolidation}\label{sec:EWC}
\noindent The approach from \cite{Kirkpatrick2016} is a typical regularization-based model for DNNs, see \cref{sec:relwork}.
EWC stores DNN parameters $\vec{\theta^{T_t}}$ after training on sub-task $T_t$.
In addition, EWC computes the \enquote{importance} of each parameter after training on sub-task $T_t$ by approximating the diagonal of the Fisher Information Matrix\footnote{See \cite{Gepperth2019SIM} for a discussion of this approximation.} $\vec F^{T_t}$. 
The EWC loss function contains additional terms, see \cref{eqn:ewc_loss} besides the cross-entropy loss computed on the current sub-task $T_c$. 
These additional terms punish deviations from \enquote{important} DNN parameter values obtained after training on past sub-tasks:
\begin{equation}
	\mathcal{L}^{EWC} = \mathcal{L}_{T_c}(\theta) + \frac{\lambda}{2}\sum_{t=1}^{c-1} \sum_i F^{T_t}_i\Big(\theta_i - \theta^{T_t}_i\Big)^2 
	\label{eqn:ewc_loss}
\end{equation} 
EWC is optimized using SGD, usually in conjunction with the Adam optimizer. 
EWC hyper-parameters comprise the SGD step size $\epsilon$, the regularization constant $\lambda$ and of course the number and size of layers in the DNN. 
In \cite{Kirkpatrick2016}, it is proposed to set $\lambda=\epsilon^{-1}$, thereby eliminating one hyper-parameter.
\section{Experimental Setup}\label{sec:exp_setup}
\noindent In this section, the used parameters, tasks and evaluation methods for EWC and GMR are described.
Generally, each experiment is repeated $10$ times.
This ensures that the results are not influenced by the random initialization of the model.
\subsection{Sequential Learning Tasks}\label{sec:SLT}
\noindent Sequential Learning Tasks (SLTs) simulate a continuous learning scenario by dividing datasets given in \cref{sec:datasets}.
The resulting sub-datasets are enumerated and contain only samples of non-overlapping classes.
For example, a $D_{5\text{-}5}$ task consists of two sub-datasets consisting of $5$ classes each.
Each sub-task is identified by its order, e.g., $T_1$, $T_2$, $\ldots$, $T_x$.
Baseline experiments ($D_{10}$) contain all available classes to investigate the effect of incremental task-by-task training.
\par
SLTs can be used to perform basic experiments to determine the effect of forgetting under the conditions described above.
To measure the impact of the number of classes contained in a task, different combinations and subdivisions are evaluated.
\Cref{tab:SLTs} shows evaluated SLTs and their definition of sub-tasks.
\begin{table}[htb]
	\centering
	\caption{Definition of Sequential Learning Tasks (SLTs) and the class divisions of their sub-tasks.
	} 
	\label{tab:SLTs}
	\setlength\tabcolsep{2pt}
	\begin{tabular}{|l|l|}
		\hline
		\textbf{SLT}                                                                                                & \textbf{Sub-tasks}                                                                                                                             \\ \hline
		$D_{10}$ (baseline)                                                                                         & $T_1($0, 1, 2, 3, 4, 5, 6, 7, 8, 9$)$                                                                                                          \\ \hline
		$D_{9\text{-}1a}$                                                                                           & $T_1($0, 1, 2, 3, 4, 5, 6, 7, 8$)$~~~$T_2($9$)$                                                                                                \\ \hline
		$D_{9\text{-}1b}$                                                                                           & $T_1($0, 1, 2, 4, 5, 6, 7, 8, 9$)$~~~$T_2($3$)$                                                                                                \\ \hline
		$D_{5\text{-}5a}$                                                                                           & $T_1($0, 1, 2, 3, 4$)$~~~$T_2($5, 6, 7, 8, 9$)$                                                                                                \\ \hline
		$D_{5\text{-}5b}$                                                                                           & $T_1($0, 1, 2, 6, 7$)$~~~$T_2($3, 4, 5, 8, 9$)$                                                                                                \\ \hline
		$D_{3\text{-}3\text{-}3\text{-}1}$                                                                          & $T_1($0, 1, 2$)$~~~$T_2($3, 4, 5$)$~~~$T_3($6, 7, 8$)$~~~$T_4($9$)$                                                                            \\ \hline
		$D_{2\text{-}2\text{-}2\text{-}2\text{-}2a}$                                                                & $T_1($0, 1$)$~~~$T_2($2, 3$)$~~~$T_3($4, 5$)$~~~$T_4($6, 7$)$~~~$T_5($8, 9$)$                                                                  \\ \hline
		$D_{2\text{-}2\text{-}2\text{-}2\text{-}2b}$                                                                & $T_1($1, 7$)$~~~$T_2($0, 2$)$~~~$T_3($6, 8$)$~~~$T_4($4, 5$)$~~~$T_5($3, 9$)$                                                                  \\ \hline
		\multirow{-2}{*}{$D_{1\text{-}1\text{-}1\text{-}1\text{-}1\text{-}1\text{-}1\text{-}1\text{-}1\text{-}1a}$} & \shortstack{$T_1($0$)$~~~$T_2($1$)$~~~$T_3($2$)$~~~$T_4($3$)$~~~$T_5($4$)$\\$T_6($5$)$~~~$T_7($6$)$~~~$T_8($7$)$~~~$T_9($8$)$~~~$T_{10}($9$)$} \\ \hline
		\multirow{-2}{*}{$D_{1\text{-}1\text{-}1\text{-}1\text{-}1\text{-}1\text{-}1\text{-}1\text{-}1\text{-}1b}$} & \shortstack{$T_1($7$)$~~~$T_2($1$)$~~~$T_3($2$)$~~~$T_4($0$)$~~~$T_5($6$)$\\$T_6($8$)$~~~$T_7($4$)$~~~$T_8($5$)$~~~$T_9($9$)$~~~$T_{10}($3$)$} \\ \hline
	\end{tabular}
\end{table}
\subsection{Hyper-Parameter}\label{sec:hyperparameter}
\noindent For both investigated models, a fixed batch size of $\mathcal{B}$\,$=$\,$100$ is used.
Both models are informed about the start and end of each sub-task, so they can react appropriately. 
\par
The following hyper-parameters were explored for GMR by performing a grid search.
GMR is trained for $50$ epochs ($\mathcal{E}$) for the first task.
The number of epochs for retraining is adjusted depending on the ratio of the number of classes of the previous and the new task.
We do not modify the model itself after each training task.
The start of a new sub-task triggers the generation of samples from past sub-tasks, see \cref{fig:gmr}.
The number of GMM components varies: $K$\,$\in$\,$\{36,84,100\}$. 
As indicated in \cite{Gepperth2019}, the principle \enquote{the more the better} applies, which makes selection unproblematic.
In order to investigate the influence of the loss function of the linear layer, we evaluate cross-entropy and MSE losses.
The learning rate of both, the GMM and the linear layer, is fixed after initial tests. 
This is reasonable due to the GMM learning rate exclusively depending on the data which do not change, and because the inputs to the linear classifier layer are normalized.
Thus, the total number of $1\,800$ experiments performed is composed of the three datasets, the conducted SLTs ($10$), varied parameters ($3$) and $10$ repetitions.
\par
For EWC, we perform a grid search for the parameter $\epsilon$.
We vary $\epsilon$ as $\epsilon$\,$\in$\,$\{0.001,0.0001,0.00001,0.000001,0.0000001\}$ while $\lambda$ is always set to $\frac{1}{\epsilon}$.
We fix the model architecture to a three-layer DNN, each of size $800$ (used in most recent studies on EWC).
Training epochs $\mathcal{E}$ are empirically set to $10$ for each training task. 
This is done to ensure a fair comparison of GMR and EWC. 
Due to replay, GMR is insensitive to the duration of training for each sub-task. 
EWC, however, is very sensible to this parameter as performance tends to deteriorate over (training) time.
As longer training times lead to less favorable EWC results, we generally evaluate performance during the training process.
Using this evaluation strategy, the high number of training iterations for EWC should not be problematic.
This contradicts a real-world scenario evaluation, since the best training duration is explicitly unknown and cannot be measured but it is used here, however, to ensure comparability in the first place.
The best hyper-parameters and experiments are selected based on the maximum measured accuracy during training averages over $10$ experiment repetitions.
From the number of datasets ($3$), the performed SLTs ($10$), the different parameters ($5$) and the repetitions ($10$) results in $1\,500$ conducted EWC experiments.
\subsection{Task Boundaries Specification}
\noindent One difficulty of continual learning (CL), which is often not addressed, is the specification of task boundaries.
This information of a task boundary alone can be used to mitigate the CF effect.
At this point, the recognition of task boundaries is not the goal of this work.
Due to an easier comparison, we specify task boundaries for both, EWC and GMR.
\par
\begin{figure}[htb]
	\centering
	\includegraphics[width=\linewidth]{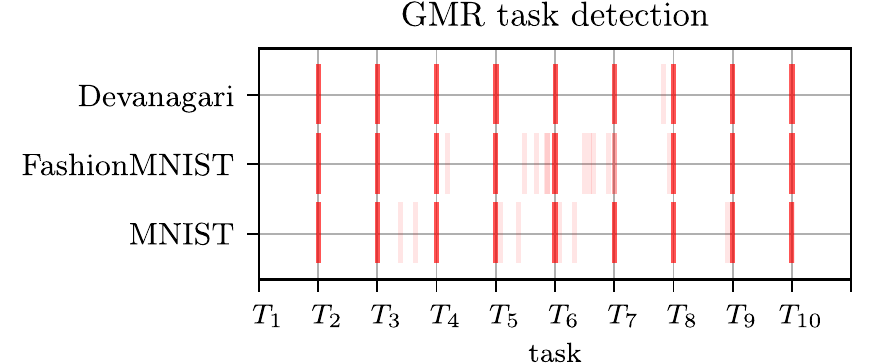}
	\caption{Visualization of GMR task detection.
			Each time a task boundary is detected, a semi-transparent line is drawn. 
			Where many experiments are certain, the boundaries become more clearly visible.
	}
	\label{fig:taskdetection}
\end{figure}

To investigate this issue, we show that GMR is able to detect task boundaries, which is illustrated by the following experiment.
We train the model class by class, one after another ($D_{1\text{-}1\text{-}1\text{-}1\text{-}1\text{-}1\text{-}1\text{-}1\text{-}1\text{-}1a}$, $T_1$, $\ldots$, $T_{10}$).
A straightforward heuristic is used, based on the log-likelihood of incoming batches of data.
A task boundary is detected if the log-likelihood is smaller than $80\%$ of the sliding log-likelihood for more than $10$ tests iterations.
Before a new task can be detected, $500$ training iterations must ave passed.
We train the model for many iterations ($\mathcal{E}$\,$=$\,$200$ each task) in order to have 1) enough time to converge at task $T_1$ and 2) to prove stability over all following tasks.
The experiment is repeated $10$ times.
The result, which can be seen in \cref{fig:taskdetection}, can be interpreted as the model properly detecting the task boundaries after already $10$\textsuperscript{th} training iterations on a new task.
\subsection{Evaluation Method}\label{sec:eval_method}
\noindent In order to compare EWC with our GMR approach, the standard classification accuracy is used.
Therefore, each Model $M$ is trained task by task.
During task training, models are tested on the test dataset of $D_{10}$ (all joint classes).
Each task is tested with the full test dataset on $10$ equally distributed test points at each task $T_t$.
For evaluation, we use the maximum accuracy measured over the full training process.
It must be kept in mind that, for example, after training the first SLT with, e.g., $5$ of $10$ ($T_1$) classes ($D_{5\text{-}5a}$), a maximum accuracy of $50\%$ on testing data ($D_{10}$) is possible.
\par
\begin{figure}[htb]
	\centering
	\includegraphics[width=\linewidth]{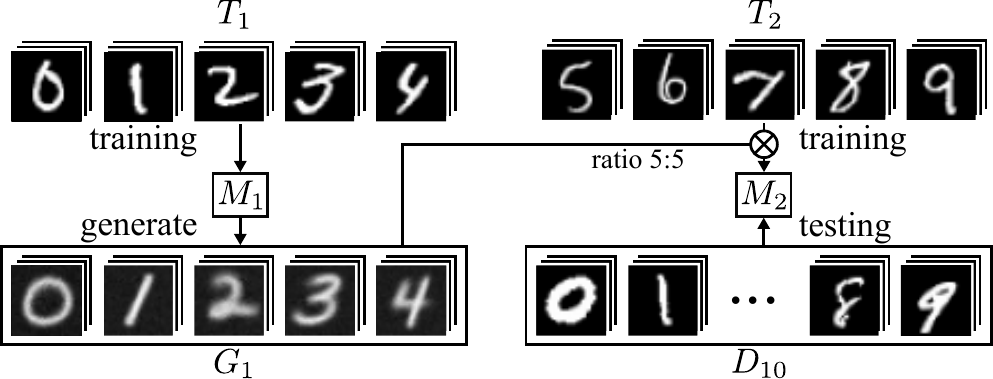}
	\caption{Visualization of the training scheme for the SLT $D_{5\text{-}5a}$ on MNIST.}
\label{fig:generative_approach}
\end{figure}

For the generative approach, examples are generated after each sub-task is completed.
For the following sub-task $T_{x+1}$ samples from $G_x$ are mixed with samples of the current task ($G_x \otimes T_{x+1}$).
To ensure that training of the following sub-task is not affected by adding generated samples, the number of training iterations must be adjusted proportionally to the quantity.
This also applies to the merging ratio of the generated and the training examples.
For example, if the number of classes is multiplied by two, the training iterations are also doubled.
These steps are repeated until the model $M_{x}$ is trained with all sub-tasks.
In \cref{fig:generative_approach} the procedure for the $D_{5\text{-}5a}$ learning task is visualized.
\section{Experimental Results}
\noindent In the following, results of the conducted $1\,800$ GMR- and $1\,500$ EWC-experiments is described.
The models \cref{sec:GMR} and \cref{sec:EWC} are used with the hyper-parameters specified in \cref{sec:hyperparameter}.
The evaluation method is introduced in \cref{sec:eval_method}.
Results of the experiment and its repetitions are expressed by the standard deviation in addition to the maximum accuracy of the baseline, respectively its difference.
Therefore, the defined SLTs of \cref{sec:SLT} are applied to the three datasets from \cref{sec:datasets}.
\par
To compare the models, each is trained at the $D_{10}$ task to obtain a baseline value.
By means of these values, the measure of forgetting is determined.
A summary of all the best experiment results is shown in \cref{tab:results}.
\begin{table*}
	\newlength{\mycolumnwidth}
	\setlength{\mycolumnwidth}{.8cm}
	\newcolumntype{b}{>{\centering\arraybackslash}m{\mycolumnwidth}}
	\centering
	\caption{Results of the conducted GMR and EWC experiments. 
             The accuracy in $\%$ of the baseline experiments on all available classes is shown for each dataset.
             For each best SLT experiment (defined in \cref{tab:SLTs}) the difference to the baseline is given.
             Therefore, the maximum measured accuracy value is used, averaged over $10$ experiment repetitions.
             To measure the accuracy, the joint test dataset consisting of all tasks ($D_{10}$) is used. 
            }
	\label{tab:results}
	\setlength\tabcolsep{5pt}
\begin{tabular}{|l|bb|bb|bb!{\vrule width 1pt}bb|bb|bb|}
	\hline
	\multirow{3}{*}{\diagbox[width=76.1pt,height=27.1pt]{SLT}{\shortstack{\textbf{model}\\dataset}}} &                                                                                   \multicolumn{6}{c!{\vrule width 1pt}}{\textbf{GMR}}                                                                                    &                                                                                             \multicolumn{6}{c|}{\textbf{EWC}}                                                                                             \\
	                                                                                                 &                       \multicolumn{2}{c|}{MNIST}                       &                   \multicolumn{2}{c|}{FashionMNIST}                    &           \multicolumn{2}{c!{\vrule width 1pt}}{Devanagari}            &                       \multicolumn{2}{c|}{MNIST}                       &                   \multicolumn{2}{c|}{FashionMNIST}                    &                     \multicolumn{2}{c|}{Devanagari}                     \\
	                                                                                                 & acc. $\%$                          & std                               & acc. $\%$                          & std                               & acc. $\%$                          & std                               & acc. $\%$                          & std                               & acc. $\%$                          & std                               & acc. $\%$                          & std                                \\ \hline
	$D_{10}$ baseline                                                                                & \tablenum[table-format=2.1]{87.4}  & \tablenum[table-format=1.2]{0.59} & \tablenum[table-format=2.1]{73.9}  & \tablenum[table-format=1.2]{0.26} & \tablenum[table-format=2.1]{74.1}  & \tablenum[table-format=1.2]{0.73} & \tablenum[table-format=2.1]{97.57} & \tablenum[table-format=1.2]{0.26} & \tablenum[table-format=2.1]{87.55} & \tablenum[table-format=1.2]{0.38} & \tablenum[table-format=2.1]{95.58} & \tablenum[table-format=2.2]{0.56}  \\ \hline
	                                                                                                 & diff.                              & std                               & diff.                              & std                               & diff.                              & std                               & diff.                              & std                               & diff.                              & std                               & diff.                              & std                                \\ \hline
	$D_{9\text{-}1a}$                                                                                & \tablenum[table-format=3.1]{-1.3}  & \tablenum[table-format=1.2]{0.59} & \tablenum[table-format=3.1]{-2.7}  & \tablenum[table-format=1.2]{0.26} & \tablenum[table-format=3.1]{-3.2}  & \tablenum[table-format=1.2]{0.73} & \tablenum[table-format=3.1]{-41.8} & \tablenum[table-format=1.2]{0.26} & \tablenum[table-format=3.1] {-9.6} & \tablenum[table-format=1.2]{0.38} & \tablenum[table-format=3.1]{-56.6} & \tablenum[table-format=2.2]{0.56}  \\ \hline
	$D_{9\text{-}1b}$                                                                                & \tablenum[table-format=3.1]{-3.5}  & \tablenum[table-format=1.2]{2.19} & \tablenum[table-format=3.1]{-1.5}  & \tablenum[table-format=1.2]{0.87} & \tablenum[table-format=3.1]{-1.4}  & \tablenum[table-format=1.2]{0.88} & \tablenum[table-format=3.1]{-50.7} & \tablenum[table-format=1.2]{7.77} & \tablenum[table-format=3.1]{-20.1} & \tablenum[table-format=1.2]{2.52} & \tablenum[table-format=3.1]{-29.7} & \tablenum[table-format=2.2]{13.34} \\ \hline
	$D_{5\text{-}5a}$                                                                                & \tablenum[table-format=3.1]{-0.6}  & \tablenum[table-format=1.2]{1.53} & \tablenum[table-format=3.1]{-1.2}  & \tablenum[table-format=1.2]{1.53} & \tablenum[table-format=3.1]{-6.8}  & \tablenum[table-format=1.2]{1.38} & \tablenum[table-format=3.1]{-35.3} & \tablenum[table-format=1.2]{6.65} & \tablenum[table-format=3.1]{-32.7} & \tablenum[table-format=1.2]{4.22} & \tablenum[table-format=3.1]{-46.0} & \tablenum[table-format=2.2]{15.38} \\ \hline
	$D_{5\text{-}5b}$                                                                                & \tablenum[table-format=3.1]{-1.3}  & \tablenum[table-format=1.2]{1.92} & \tablenum[table-format=3.1]{-1.9}  & \tablenum[table-format=1.2]{0.49} & \tablenum[table-format=3.1]{-4.7}  & \tablenum[table-format=1.2]{1.59} & \tablenum[table-format=3.1]{-35.0} & \tablenum[table-format=1.2]{1.83} & \tablenum[table-format=3.1]{-36.0} & \tablenum[table-format=1.2]{2.72} & \tablenum[table-format=3.1]{-47.1} & \tablenum[table-format=2.2]{0.11}  \\ \hline
	$D_{3\text{-}3\text{-}3\text{-}1}$                                                               & \tablenum[table-format=3.1]{-8.3}  & \tablenum[table-format=1.2]{2.68} & \tablenum[table-format=3.1]{-8.3}  & \tablenum[table-format=1.2]{1.46} & \tablenum[table-format=3.1]{-15.5} & \tablenum[table-format=1.2]{1.83} & \tablenum[table-format=3.1]{-59.4} & \tablenum[table-format=1.2]{2.99} & \tablenum[table-format=3.1]{-52.7} & \tablenum[table-format=1.2]{2.57} & \tablenum[table-format=3.1]{-63.9} & \tablenum[table-format=2.2]{0.11}  \\ \hline
	$D_{2\text{-}2\text{-}2\text{-}2\text{-}2a}$                                                     & \tablenum[table-format=3.1]{-9.5}  & \tablenum[table-format=1.2]{3.83} & \tablenum[table-format=3.1]{-8.5}  & \tablenum[table-format=1.2]{0.91} & \tablenum[table-format=3.1]{-22.5} & \tablenum[table-format=1.2]{2.71} & \tablenum[table-format=3.1]{-72.2} & \tablenum[table-format=1.2]{7.43} & \tablenum[table-format=3.1]{-55.6} & \tablenum[table-format=1.2]{4.05} & \tablenum[table-format=3.1]{-72.1} & \tablenum[table-format=2.2]{2.75}  \\ \hline
	$D_{2\text{-}2\text{-}2\text{-}2\text{-}2b}$                                                     & \tablenum[table-format=3.1]{-10.4} & \tablenum[table-format=1.2]{5.28} & \tablenum[table-format=3.1]{-5.7}  & \tablenum[table-format=1.2]{2.37} & \tablenum[table-format=3.1]{-14.7} & \tablenum[table-format=1.2]{2.94} & \tablenum[table-format=3.1]{-72.6} & \tablenum[table-format=1.2]{3.22} & \tablenum[table-format=3.1]{-57.3} & \tablenum[table-format=1.2]{4.99} & \tablenum[table-format=3.1]{-73.2} & \tablenum[table-format=2.2]{2.31}  \\ \hline
	$D_{1\text{-}1\text{-}1\text{-}1\text{-}1\text{-}1\text{-}1\text{-}1\text{-}1\text{-}1a}$        & \tablenum[table-format=3.1]{-23.3} & \tablenum[table-format=1.2]{4.10} & \tablenum[table-format=3.1]{-19.2} & \tablenum[table-format=1.2]{1.83} & \tablenum[table-format=3.1]{-27.5} & \tablenum[table-format=1.2]{7.52} & \tablenum[table-format=3.1]{-74.5} & \tablenum[table-format=1.2]{3.79} & \tablenum[table-format=3.1]{-55.2} & \tablenum[table-format=1.2]{5.64} & \tablenum[table-format=3.1]{-71.1} & \tablenum[table-format=2.2]{2.26}  \\ \hline
	$D_{1\text{-}1\text{-}1\text{-}1\text{-}1\text{-}1\text{-}1\text{-}1\text{-}1\text{-}1b}$        & \tablenum[table-format=3.1]{-16.3} & \tablenum[table-format=1.2]{2.06} & \tablenum[table-format=3.1]{-19.6} & \tablenum[table-format=1.2]{2.06} & \tablenum[table-format=3.1]{-35.6} & \tablenum[table-format=1.2]{2.69} & \tablenum[table-format=3.1]{-76.6} & \tablenum[table-format=1.2]{3.40} & \tablenum[table-format=3.1]{-56.0} & \tablenum[table-format=1.2]{4.17} & \tablenum[table-format=3.1]{-77.2} & \tablenum[table-format=2.2]{2.74}  \\ \hline
\end{tabular} 	
\end{table*}
\par
\begin{figure}[htb]
	\centering
	\includegraphics[width=\linewidth]{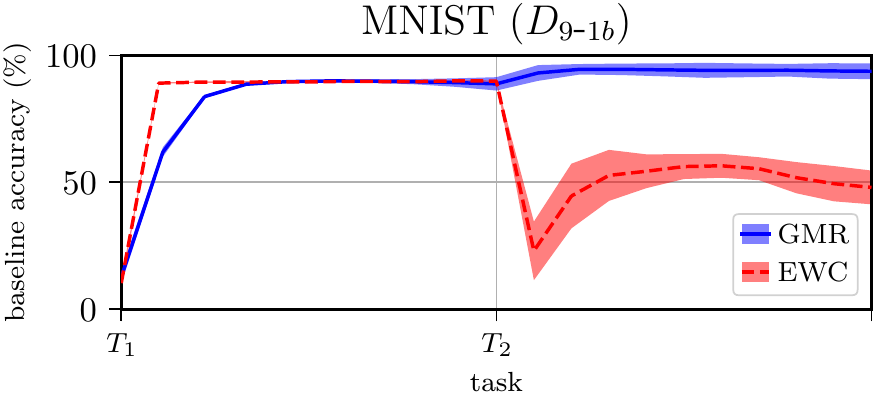}
	\caption{Visualization of accuracy trend for SLT $D_{9\text{-}1b}$.}
	\label{fig:mnistd9-1a}
\end{figure}

We visualize the measured accuracy trend to give deeper insights into the comparison of the models.
Here, we only show results for the MNIST dataset.
For the other datasets, the trends look similar, they just slightly shifted (see \cref{tab:results}).
\Cref{fig:mnistd9-1a} shows the trend for SLT $D_{9\text{-}1a}$ for the best GMR and EWC experiments.
The faces indicate the standard deviation of the accuracy over $10$ experiment repetitions.
The lines show the average of the accuracy across the repetitions.
The specified \textit{baseline accuracy} in the diagrams shows the normalized accuracy based on the best $D_{10}$ baseline experiments of each model.
What becomes clear in \cref{fig:mnistd9-1a} is the steeply falling slope of the EWC experiment starting with training on $T_2$.
At the same time, the accuracy of GMR increases slightly by training on only one additional class.
It also shows that it is very consistent until the end.
This effect is not recognizable with EWC.
On the contrary, the longer the training, the more knowledge is forgotten.
\par
\begin{figure}[htb!]
	\centering
	\includegraphics[width=\linewidth]{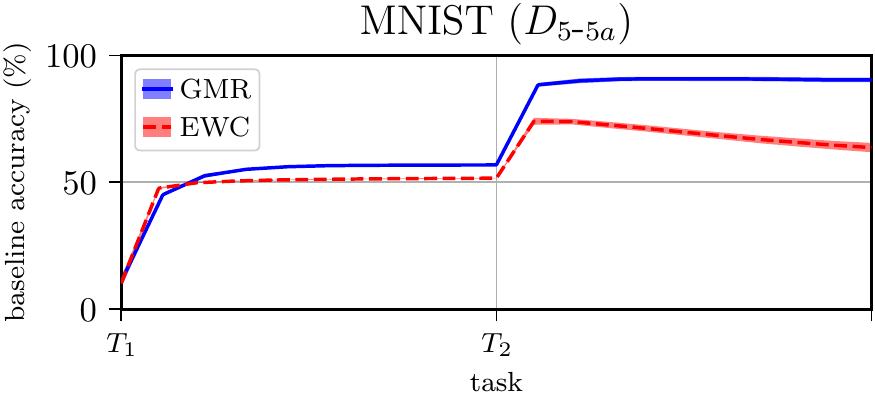}
	\caption{Visualization of accuracy trend for SLT $D_{5\text{-}5a}$.}
	\label{fig:mnistd5-5}
\end{figure}
\begin{figure}[htb!]
	\centering
	\begin{subfigure}[c]{0.49\linewidth}
	\includegraphics[width=\linewidth,trim=1.9cm 0cm 1.9cm 0cm,clip]{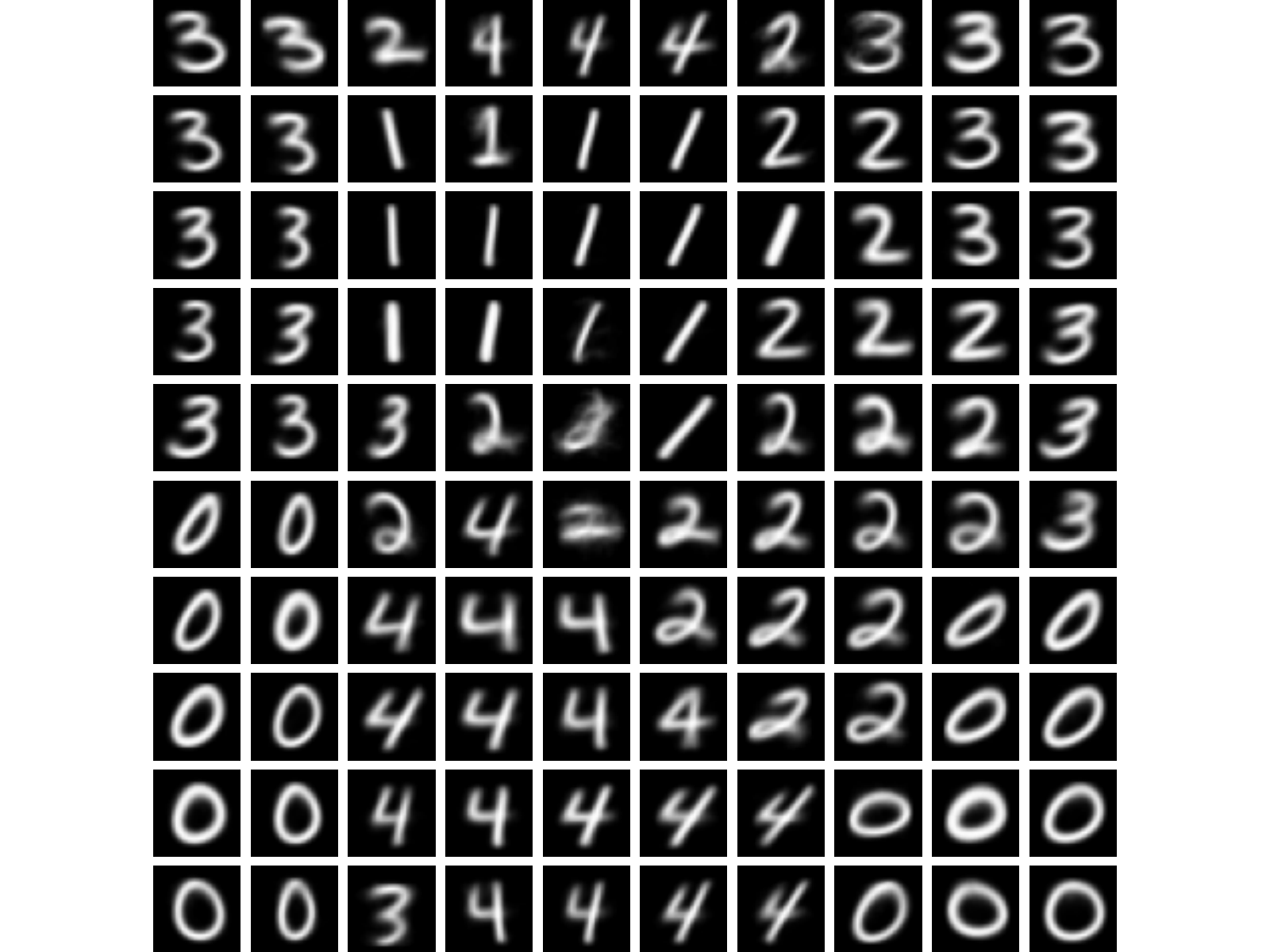}
	\subcaption{Prototypes ($\mu$) after training $T_1$.\label{fig:T1_protos}}
	\end{subfigure}
	\begin{subfigure}[c]{0.49\linewidth}
		\includegraphics[width=\linewidth,trim=1.9cm 0cm 1.9cm 0cm,clip]{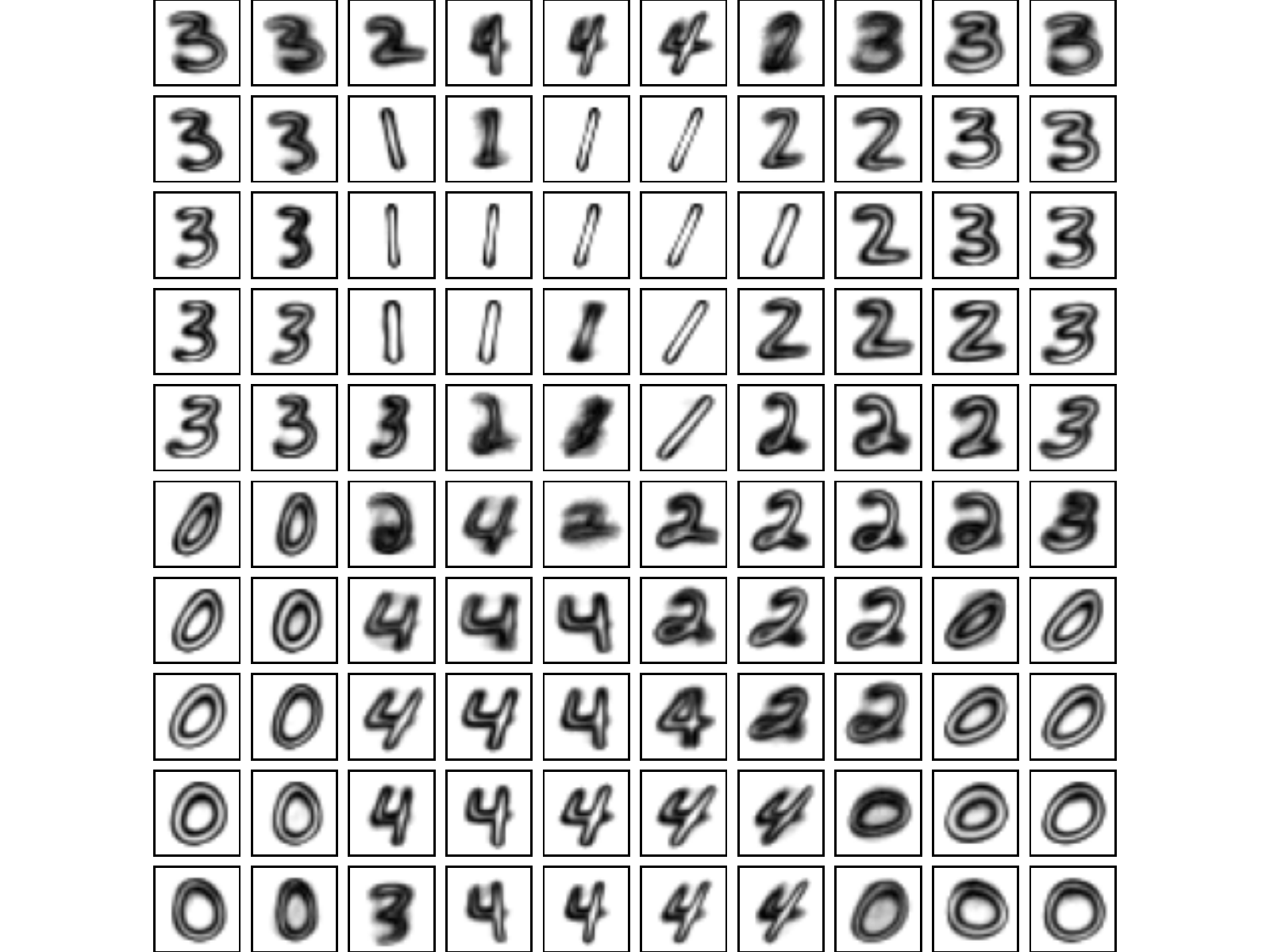}
		\subcaption{Variances ($\Sigma$) after training $T_1$.\label{fig:T1_variance}}
	\end{subfigure}
	\\\vspace{1em}
	\begin{subfigure}[c]{0.49\linewidth}
		\includegraphics[width=\linewidth,trim=1.9cm 0cm 1.9cm 0cm,clip]{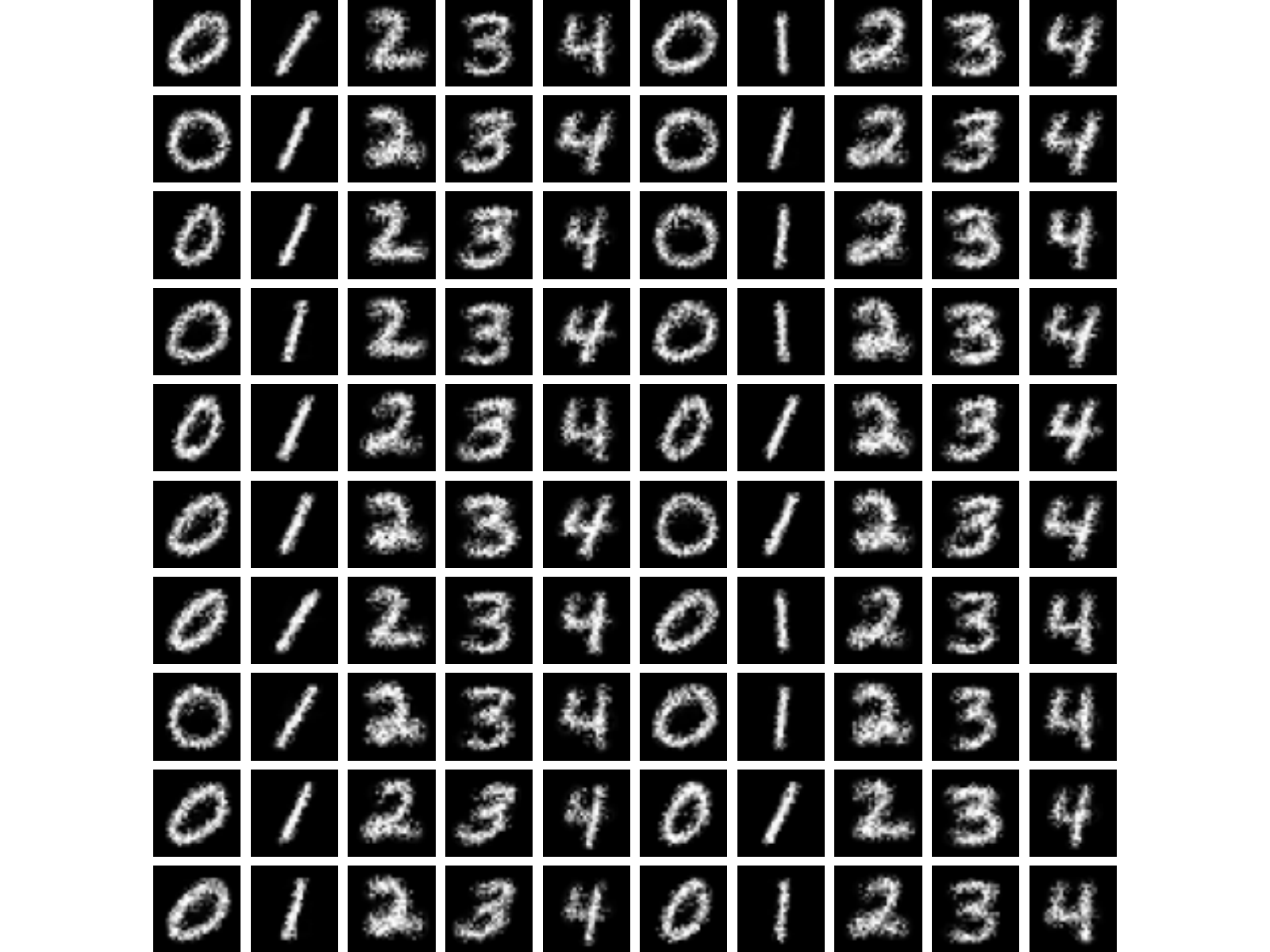}
		\subcaption{Batch of generated samples ($G_1$).\label{fig:T1_samples}}
	\end{subfigure}
	\\\vspace{1em}
	\begin{subfigure}[c]{0.49\linewidth}
		\includegraphics[width=\linewidth,trim=1.9cm 0cm 1.9cm 0cm,clip]{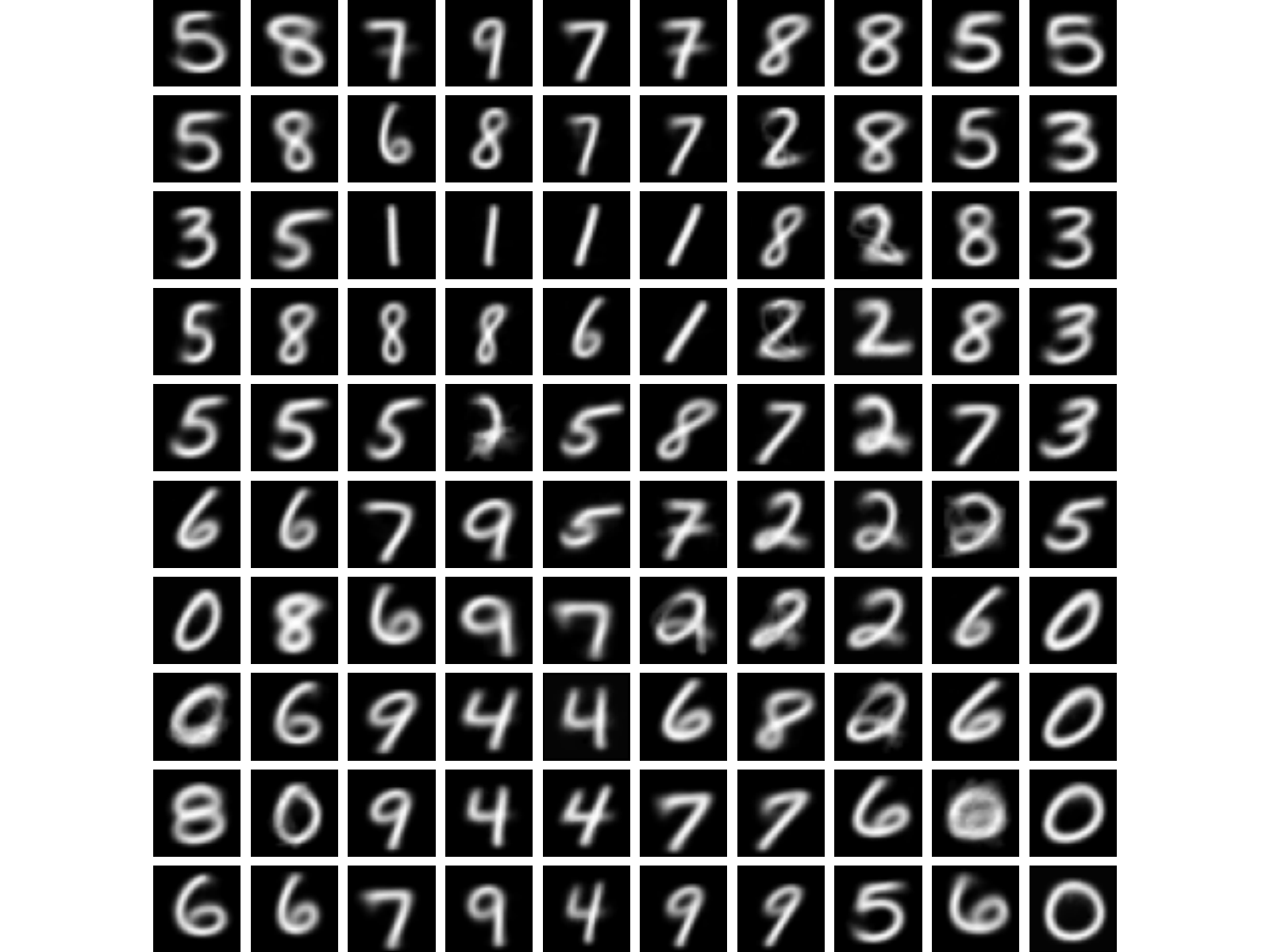}
		\subcaption{Prototypes after training $G_1$\,$\otimes$\,$T_1$.\label{fig:T2_protos}}
	\end{subfigure}
	\begin{subfigure}[c]{0.49\linewidth}
		\includegraphics[width=\linewidth,trim=1.9cm 0cm 1.9cm 0cm,clip]{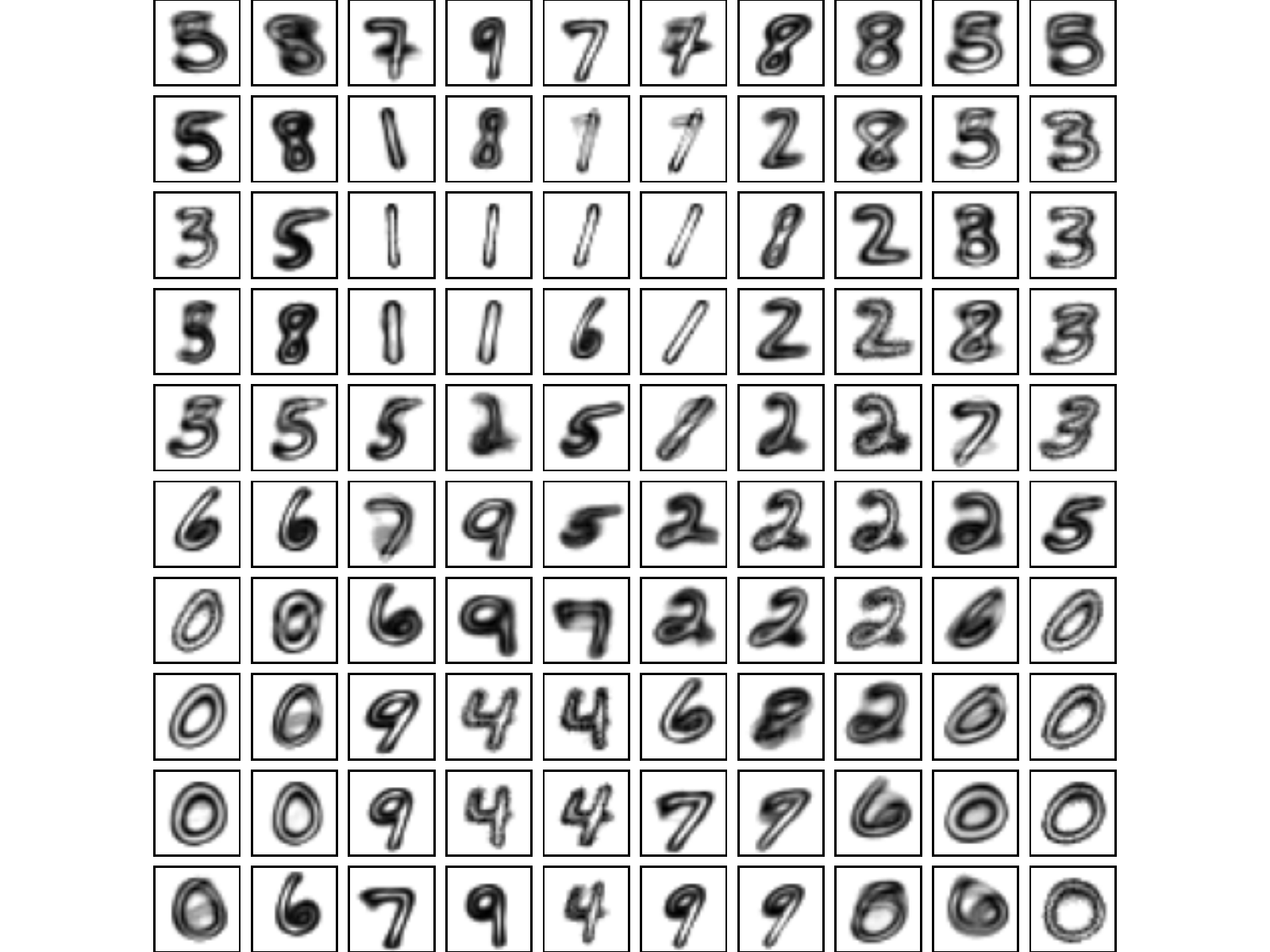}
		\subcaption{Variances after training $G_1$\,$\otimes$\,$T_2$.\label{fig:T2_variance}}
	\end{subfigure}
	\caption{Visualization of GMM prototype, GMM variances and generated samples for SLT $D_{5\text{-}5a}$: $T_1($0,1,2,3,4$)$ and $T_2($5,6,7,8,9$)$.}
	\label{fig:examples}
\end{figure}
\begin{figure}[htb!]
	\centering
	\includegraphics[width=\linewidth]{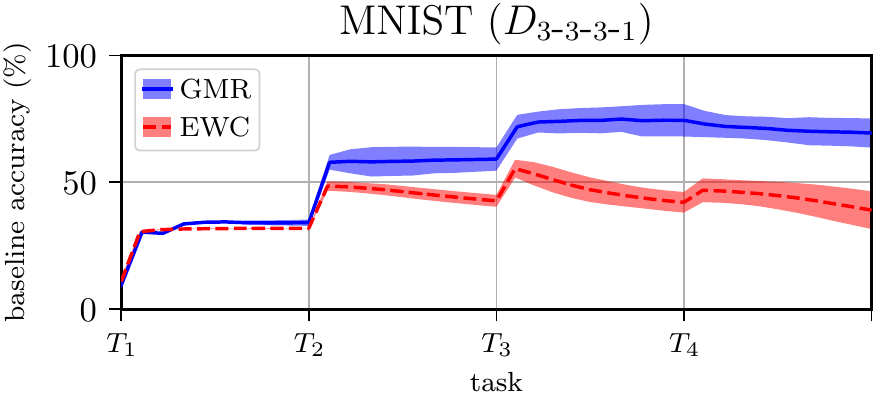}
	\caption{Visualization of accuracy trend for SLT $D_{3\text{-}3\text{-}3\text{-}1}$.}
	\label{fig:mnistd3-3-3-1}
\end{figure}
\begin{figure}[htb!]
	\centering	
	\includegraphics[width=\linewidth]{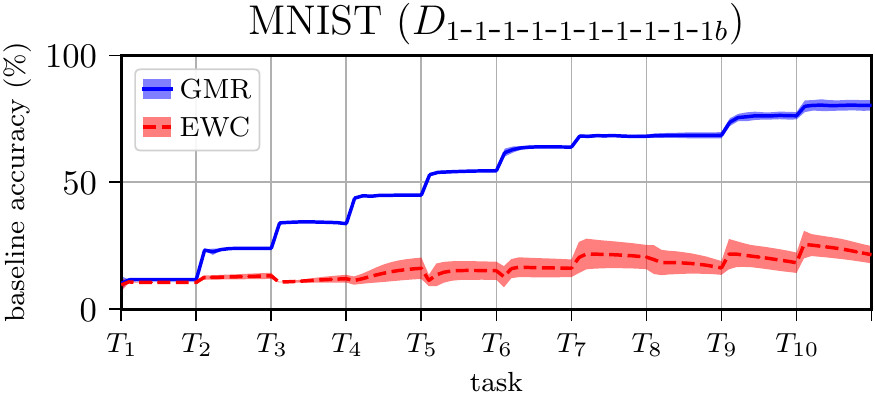} 
	\caption{Visualization of accuracy trend for SLT $D_{1\text{-}1\text{-}1\text{-}1\text{-}1\text{-}1\text{-}1\text{-}1\text{-}1\text{-}1b}$.}  
	\label{fig:mnistd1-1-1-1-1-1-1-1-1-1b} 
\end{figure}

This effect of forgetting can be recognized in all EWC experiments (see also \cref{fig:mnistd5-5,fig:mnistd3-3-3-1,fig:mnistd1-1-1-1-1-1-1-1-1-1b}).
For the best $D_{5\text{-}5a}$ experiment (see \cref{fig:mnistd5-5}) we want to give more insights into the (re-)training process, especially the GMM layer.
After training $T_1$ for $50$ epochs with the MNIST dataset, the $100$ components $K$ respectively the means/prototypes ($\vec{\mu}$) are visualized in \cref{fig:T1_protos}.
The corresponding variances are displayed in \cref{fig:T1_variance}.
Note that the topological order of the prototypes caused by the initial annealing process is no longer preserved after $T_2$.
After finishing the training on $T_1$ a dataset $G_1$ is generated which is used for replay (see \cref{fig:T1_samples}).
The examples shown here are noisy, which is due to the high variances present in the original data.
After training with $G_1$\,$\otimes$\,$T_2$ the prototypes and variances can be observed in \cref{fig:T2_protos} and \cref{fig:T2_variance}.
This is the only experiment for which the cross-entropy loss for the linear layer give the best results.
It can be seen that the use of MSE objective function leads to a higher variance within the other GMR experiment groups.


\section{Discussion}
\noindent The experimental results underline that GMR shows consistently better continual learning performance on all tested datasets, for all sequential learning tasks. 
This effect is remarkable since GMR is significantly less complex in terms of free model parameters and memory requirements.
In addition, GMR has been shown to be capable of detecting task boundaries.
While GMR and EWC are informed about sub-task boundaries in the current training setup, this is not really a requirement for GMR.
Moreover, EWC has no possibility to determine this, as DNNs are known to be incapable of out-of-sample recognition or outlier detection. 
A similar argument distinguishes GMR from generative replay approaches as discussed in \cite{Shin2017}. 
Furthermore, we find that GMR can be re-trained with a constant number of current and generated samples and, thus, has constant time complexity w.r.t.\ the number of previous sub-tasks.
Lastly, GMR is a compact model in comparison to other replay approaches such as \cite {Shin2017}, because it integrates learner and generator in a single structure.
\par
\noindent\textbf{Baseline performance} is the first issue that needs to be discussed.
Due to GMR's simple structure, baseline (i.e., non-continual) performance on MNIST is only $87.4\%$, as compared to EWC ($97.5\%$).
First of all, it is possible to overcome this limitation by including a \textit{deep} GMM into GMR, which boosts baseline performance to about $97\%$ for a two-layer convolutional GMM (this will be discussed in a separate article).
More importantly, this article is about continual learning, and about continual learning performance. 
The results of \cref{tab:results} plainly show that GMR is superior to EWC by a large margin, even though baseline performance is rather weak. 
\par
\noindent\textbf{Suitability for applications} is another important point (see \cite{Pfuelb2019}) in which GMR surpasses EWC.
While the only real free parameter is the number of GMM components $K$, which can be set according to a \textit{the-more-the-better} strategy without reference to data.
On the contrary, the EWC parameter $\lambda$ has a strong impact on performance and must be carefully adapted to the problem by cross-validation. 
This requires using data from all sub-tasks, which contradicts the principle of continuous learning.
This includes that the sub-tasks are processed sequentially. 
\section{Future Work}
\noindent Future work will address the following issues:
\par
\noindent\textbf{Weak baseline performance} will be resolved using deep convolutional GMMs, consisting of multiple GMM and folding layers.
Since this improves the quality of sampling, a further increase in CL performance can be expected.
\par
\noindent\textbf{More refined replay strategies} might include strategies in which the model itself decides which and how many data samples to generate.
A guiding principle could be to generate samples whose internal representation would otherwise be overwritten by current sub-task data.
In comparison, this task is easier for GMMs than for DNNs.
\par
\noindent\textbf{Systematic comparison to other CL models} should be a logical next step once a sufficiently high baseline performance can be reported. 
In particular, a comparison to generative replay models like \cite{Shin2017} could be interesting.
{\small
	\bibliographystyle{ieee_fullname}
	\bibliography{ijcnn2021-replay}
}
\end{document}